\documentclass{article}

\usepackage{microtype}
\usepackage{graphicx}
\usepackage{subcaption}
\usepackage{booktabs} 
\usepackage{makecell}
\usepackage[table]{xcolor}
\usepackage{color}
\usepackage{pifont} 
\newcommand{\cmark}{\ding{51}\xspace}%
\newcommand{\xmarkg}{\textcolor{lightgray}{\ding{55}}\xspace}%

\usepackage{hyperref}

\usepackage[accepted]{icml2026}
\usepackage{marvosym}

\usepackage{amsmath}
\usepackage{amssymb}
\usepackage{mathtools}
\usepackage{amsthm}

\usepackage[capitalize,noabbrev]{cleveref}

\usepackage{caption}
\usepackage{xspace}
\usepackage{multirow}
\usepackage{tabularx}
\usepackage{amssymb}
\usepackage{makecell} 
\newcolumntype{Y}{>{\centering\arraybackslash}X}

\newcommand{\methodname}{OcclusionFormer\xspace}
\newcommand{\datasetname}{SA-Z\xspace}

\theoremstyle{plain}

\theoremstyle{definition}

\theoremstyle{remark}

\usepackage[textsize=tiny]{todonotes}

\icmltitlerunning{OcclusionFormer: Arranging Z-Order for Layout-Grounded Image Generation}

\begin{document}

\twocolumn[
  \icmltitle{\methodname: Arranging Z-Order for Layout-Grounded Image Generation}
    \vspace{-0.2cm}


  \icmlsetsymbol{equal}{*}
    \icmlsetsymbol{corr}{\raisebox{0pt}{\small \Letter}}
  \vspace{-0.1cm}
  \begin{icmlauthorlist}
    \icmlauthor{Ziye Li}{Fudan}
    \icmlauthor{Henghui Ding}{Fudan,corr}
  \end{icmlauthorlist}

  \icmlaffiliation{Fudan}{Institute of Big Data, College of Computer Science and Artificial Intelligence, Fudan University, China}

  \icmlcorrespondingauthor{Henghui Ding}{henghui.ding@gmail.com}
  
\begin{center}
    \textbf{Project Page:} \url{https://henghuiding.com/OcclusionFormer/}
    \vspace{-0.4cm}
\end{center}

  \icmlkeywords{Machine Learning, ICML}

  \vskip 0.3in

    \begin{center} 
    \centering 
    \captionsetup{type=figure}

    {\includegraphics[width=\textwidth]{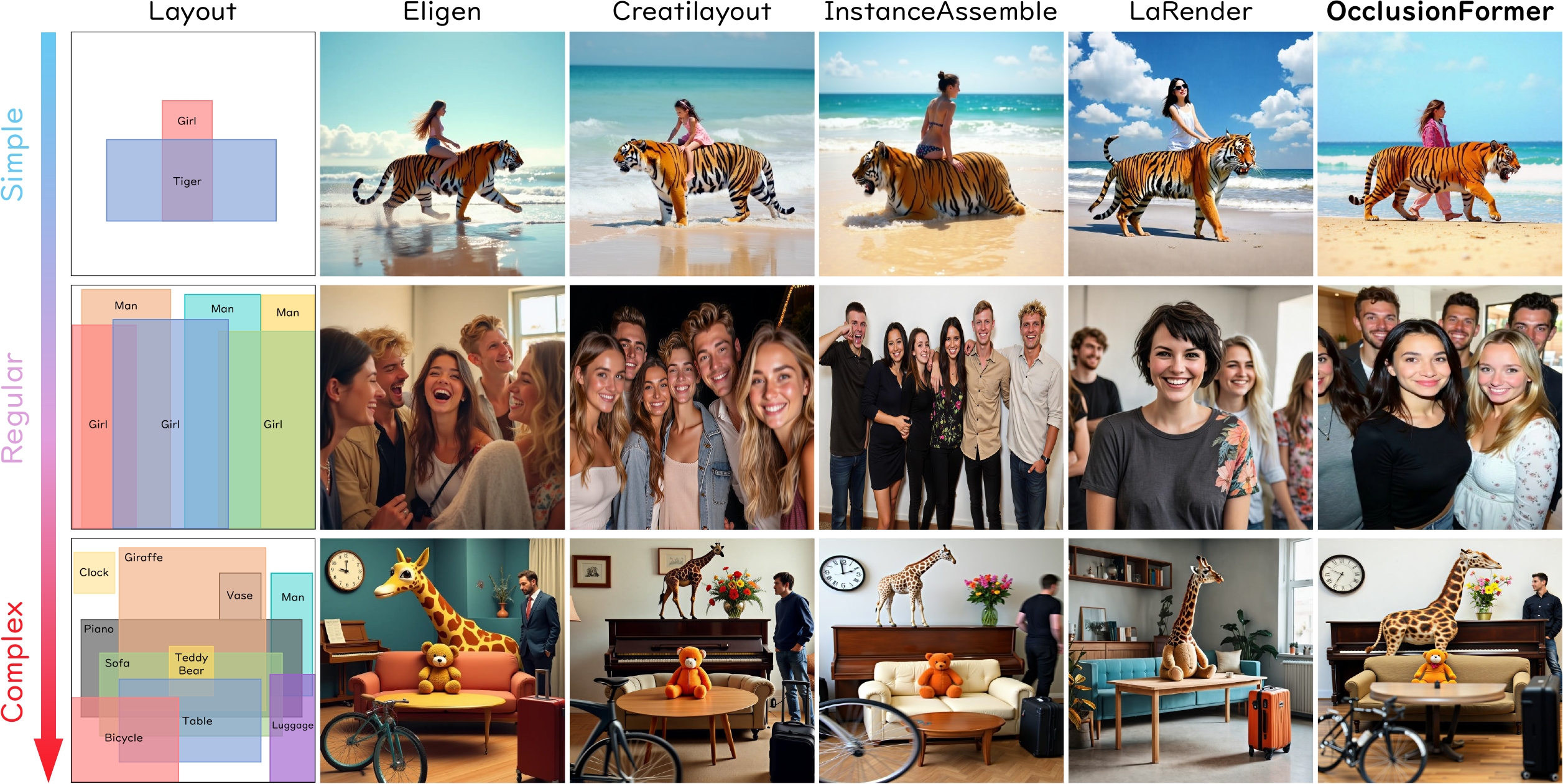}}

    \captionof{figure}{Comparison with state-of-the-art methods. 
    The first column illustrates the layout condition with multiple bounding boxes and occlusion ordering (Z-order), where foreground boxes partially occlude background ones.
    The results demonstrate that the proposed \methodname consistently outperforms prior methods under both simple and complex overlap patterns.}
    
    \label{fig:teaser}
  \end{center}

  \vskip 0.16in
  \vspace{-0.1cm}
]



\vspace{-0.1cm}
\printAffiliationsAndNotice{}  

\begin{abstract}
\vspace{-0.1cm}
  Recent layout-to-image models have achieved remarkable progress in spatial controllability. However, they still struggle with inter-object occlusion. When bounding boxes overlap, most existing methods lack explicit occlusion information, which makes the generation in intersection regions inherently ambiguous and hinders the determination of complex occlusion relationships. As a result, they often produce entangled textures or physically inconsistent layering in the overlapped areas. To address this issue, we first construct \textbf{\textit{\datasetname}}, a large-scale dataset enriched with explicit occlusion ordering and pixel-level annotations. Building upon our proposed dataset, we introduce \textbf{\textit{\methodname}}, a novel occlusion-aware Diffusion Transformer framework that explicitly models Z-order priority by decoupling instances and compositing them via volume rendering. Furthermore, to ensure fine-grained spatial precision, we introduce a queried alignment loss that explicitly supervises individual instances and enhances semantic consistency. The proposed method effectively reduces ambiguity in overlapping regions, enforces correct occlusion dependencies, and preserves structural integrity, leading to substantial accuracy gains across diverse scenes.
\end{abstract}

\vspace{-0.75cm}

\section{Introduction} 

Layout-to-image generation~\cite{li2023gligen} extends text-conditioned image generation by introducing explicit layout constraints, enabling finer-grained spatial controllability.
By leveraging 2D/3D bounding boxes~\cite{zhang2023adding,li2023gligen,zhou2024migc,wang2024instancediffusion,cheng2024hico,zhang2025eligen, zhang2025creatilayout,xiang2025instanceassemble,qin2025scenedesigner,he2025plangen} or image signals~\cite{lv2024place,liseg2any,li2025anyi2v,sun2024anycontrol,chen2024region,lin2024ctrl,mo2024freecontrol} as spatial guidance, these methods allow users to specify object locations and scales with high precision. Such capability is important for applications requiring strong structural fidelity, such as complex scene composition and visual storytelling, where the intended spatial arrangement must be faithfully preserved.

However, most existing methods largely overlook the challenge of inter-object occlusion. Unlike computer graphics pipelines that use a Z-buffer to resolve occlusion, they lack an explicit \textit{Z-order} that specifies the depth priority determining occlusion. While effective for isolated instances, these methods struggle with overlaps where intersecting boxes create ambiguity. Rather than resolving occlusion, they typically treat overlaps as feature mixtures, without explicitly distinguishing spatially overlapped instances. This can lead to entangled textures and physically inconsistent layering in the intersecting areas, ultimately harming visual realism.

This limitation also conflicts with the intuitive user workflow. As shown in \cref{fig:teaser}, users naturally provide amodal bounding boxes that specify the full object extent regardless of occlusion, rather than delineating only visible fragments. They then expect the model to follow their intended Z-order to resolve inter-object interactions. However, without explicit Z-order modeling, existing methods often misinterpret overlaps as conflicting spatial conditions and force objects to shrink into the visible area or merge unnaturally. These artifacts ultimately violate the user’s compositional intent.

A notable attempt to address this issue is LaRender~\cite{zhan2025larender}, which simulates occlusion via training-free volumetric rendering~\cite{mildenhall2020nerf}. However, it repurposes the cross-attention space in the diffusion model for occlusion control, which prevents the use of global prompts. Furthermore,
its heuristic latent manipulation is sensitive to hyperparameter choices, compromising spatial precision. 
As shown in \cref{fig:teaser}, LaRender may deviate from the specified layout under heavy overlaps, and its performance can drop in complex scenes where unsupervised guidance struggles to resolve complex occlusion dependencies.

To bridge this gap, we contend that data-driven explicit supervision is essential. We first construct \textbf{\textit{\datasetname}}, a large-scale dataset enriched with detailed pixel-level captions and explicit Z-order annotations. Additionally, we leverage SAM-3D~\cite{chen2025sam} to reconstruct 3D geometry and derive amodal annotations for occluded instances. Building on this foundation, we propose \textbf{\textit{\methodname}}, a novel framework that learns to explicitly model Z-axis priority. By integrating volumetric rendering with instance decoupling, our approach resolves depth dependencies via transmittance calculation, ensuring correct occlusion. Unlike previous heuristics, our approach maintains high fidelity even in challenging scenarios. Finally, while OverLayBench~\cite{li2025overlaybench} serves as a valuable benchmark centered on occlusion, it relies on synthetic images. To address this domain gap, we curate a challenging real-world benchmark from our \datasetname to serve as a rigorous testbed for complex occlusion. Our main contributions are summarized as follows:
\vspace{-0.305cm}
\begin{itemize}
    \item We introduce \textit{\datasetname}, a large-scale dataset enriched with detailed pixel-level instance captions and explicit Z-order annotations, and we further employ SAM-3D to derive amodal annotations via 3D reconstruction.
    \vspace{-0.305cm}
    \item We propose \textit{\methodname}, an occlusion-aware framework based on DiT that explicitly models Z-order priority. It decouples the components first, then utilizes volumetric rendering for occlusion dependencies and a queried alignment loss for supervising individual instances and enhancing semantic consistency.
    \vspace{-0.305cm}
    \item Extensive experiments demonstrate that our method establishes a new state-of-the-art in the area of occlusion control, outperforming existing baselines in resolving complex occlusion and preserving semantic integrity.
\end{itemize}

\section{Related Works}
\subsection{Layout-to-Image Generation}
\textbf{Training-free Methods.} Training-free approaches~\cite{xie2023boxdiff, bar2023multidiffusion,li2025control} enforce spatial constraints at inference time by manipulating attention maps. LaRender~\cite{zhan2025larender} further introduces volumetric rendering principles to simulate occlusion control. However, since these methods depend on heuristic gradients or latent edits instead of learned priors, they are often unstable and highly sensitive to hyperparameters. Thus, even with overlap handling, LaRender often fails to keep accurate spatial control in complex, multi-instance scenes.

\noindent\textbf{Training-based Methods.} Training-based methods inject stronger spatial guidance by adding trainable modules to diffusion backbones. Works based on U-Net such as GLIGEN~\cite{li2023gligen} and DiT-based models including Eligen~\cite{zhang2025eligen} and Creatilayout~\cite{zhang2025creatilayout} fuse box coordinates with visual features, typically improving fidelity and stability over training-free baselines. Nevertheless, they usually encode layout as a flattened 2D condition and overlook inter-object occlusion. Without an explicit occlusion-ordering mechanism, overlapping boxes yield ambiguous condition, causing feature entanglement where object appearances are unnaturally mixed.

\begin{table*}[t]
\begin{center}
\footnotesize
\caption{\textbf{Statistical comparison of datasets.} \datasetname\ features high-resolution, open-vocabulary, and 3D-aware annotations with rich geometric constraints compared to prior art. 
$\dagger$: Resolution is classified as High if the image's long edge $>1000$px.
$\ddagger$: SACap-1M derives captions from bounding box crops, where boxes often encompass irrelevant instances, this introduces visual noise into the generated texts.}
\label{tab:dataset_comparison}
\setlength{\tabcolsep}{3.75pt}
\resizebox{0.996\textwidth}{!}{\begin{tabular}{lcccccccccc}
\toprule
\textbf{Dataset} & \textbf{Source} & \textbf{\#Image} & \textbf{\#Instance} & \textbf{Resolution$^\dagger$} & \textbf{Vocabulary} & \textbf{Instance Caption} & \textbf{BBox} & \textbf{Mask} & \textbf{Z-order} & \textbf{Amodal} \\
\midrule
COCO 2017~\cite{lin2014microsoft} & COCO & $\approx$ 0.1M & $\approx$ 0.88M & Low & 80 & Class & \cmark & \cmark & \xmarkg & \xmarkg \\
InstaOrder~\cite{lee2022instance} & COCO & $\approx$ 0.1M & $\approx$ 0.50M & Low & 80 & Class & \cmark & \cmark & \cmark & \xmarkg \\
COCOA~\cite{zhu2017semantic} & COCO & $\approx$ 5.5K & $\approx$ 69.0K & Low & 80 & Class & \cmark & \cmark & \cmark & \cmark \\
OpenImages~\cite{kuznetsova2020open} & Flickr & $\approx$ 1.9M & $\approx$ 16.0M & Vary & 600 & Class & \cmark & \cmark & \xmarkg & \xmarkg \\
Visual Genome~\cite{krishna2017visual} & VG & $\approx$ 0.1M & $\approx$ 3.84M & Low & 33,877 &  Phrase & \cmark & \xmarkg & \xmarkg & \xmarkg \\
\midrule
Eligen-Data~\cite{zhang2025eligen} & Eligen & $\approx$ 0.5M & $\approx$ 1.26M & High & Open & Phrase & \cmark & \xmarkg & \xmarkg & \xmarkg \\
LayoutSAM~\cite{zhang2025creatilayout} & SA-1B & $\approx$ 2.0M & $\approx$ 10.7M & High & Open & Phrase & \cmark & \xmarkg & \xmarkg & \xmarkg \\
SACap-1M~\cite{liseg2any} & SA-1B & $\approx$ 1.0M & $\approx$ 5.88M & High & Open & \ \ Phrase$^\ddagger$ & \cmark & \cmark & \xmarkg & \xmarkg \\
\midrule
\rowcolor{cyan!10}\textbf{\datasetname} (Ours) & SA-1B & $\approx$ 1.0M & $\approx$ 5.69M & High & Open & Phrase & \cmark & \cmark & \cmark & \cmark \\
\bottomrule
\end{tabular}}
\end{center}
\vspace{-0.2in}
\end{table*}

\subsection{Datasets for Layout-to-Image Generation}
High-quality annotations are essential for training Layout-to-Image models with precise control. Early efforts used COCO~\cite{lin2014microsoft} but were limited in scale. Recent datasets such as Eligen-Data~\cite{zhang2025eligen}, LayoutSAM~\cite{zhang2025creatilayout}, and SACap-1M~\cite{liseg2any} have significantly expanded data volume and annotation richness. 
However, they remain in 2D plane, overlooking Z-axis occlusion and invisible parts of objects that are vital for handling dense layouts.
While specialized datasets like COCOA~\cite{zhu2017semantic} and InstaOrder~\cite{lee2022instance} provide Z-orders or amodal masks, they are inherently constrained by the low resolution and closed-set vocabulary of their underlying COCO images, rendering them unsuitable for modern open-vocabulary generation. To bridge this gap, we propose \textit{\datasetname}, adapted from SACap-1M. We refine the dataset by generating pixel-level captions via DescribeAnything~\cite{lian2025describe}, predicting pairwise instance Z-orders with InstaOrderNet~\cite{lee2022instance}, and estimating amodal annotations using SAM-3D~\cite{chen2025sam}. Detailed statistics of \datasetname are provided in~\cref{tab:dataset_comparison}.
\vspace{-0.1cm}

\section{Method}

\subsection{Preliminaries}
\label{sec:preliminaries}

\paragraph{Volume Rendering.}
Volumetric rendering~\cite{mildenhall2020nerf} is a differentiable mechanism that aggregates features along a ray $\mathbf{r}$ via integral accumulation:
\begin{equation}
    \hat{\mathbf{C}}(\mathbf{r}) = \sum_{i=1}^{N} T_i \alpha_i \mathbf{c}_i,
\end{equation}
where $\mathbf{c}_i$ is the feature at step $i$, and $\alpha_i = 1 - \exp(-\sigma_i)$ is opacity derived from density $\sigma_i$. $T_i = \exp(-\sum_{j=1}^{i-1} \sigma_j)$ denotes the transmittance, representing the probability of the ray remaining unblocked up to the step $i$ . 
\vspace{-0.2cm}

\begin{figure}
\centering
\includegraphics[width=1\linewidth]{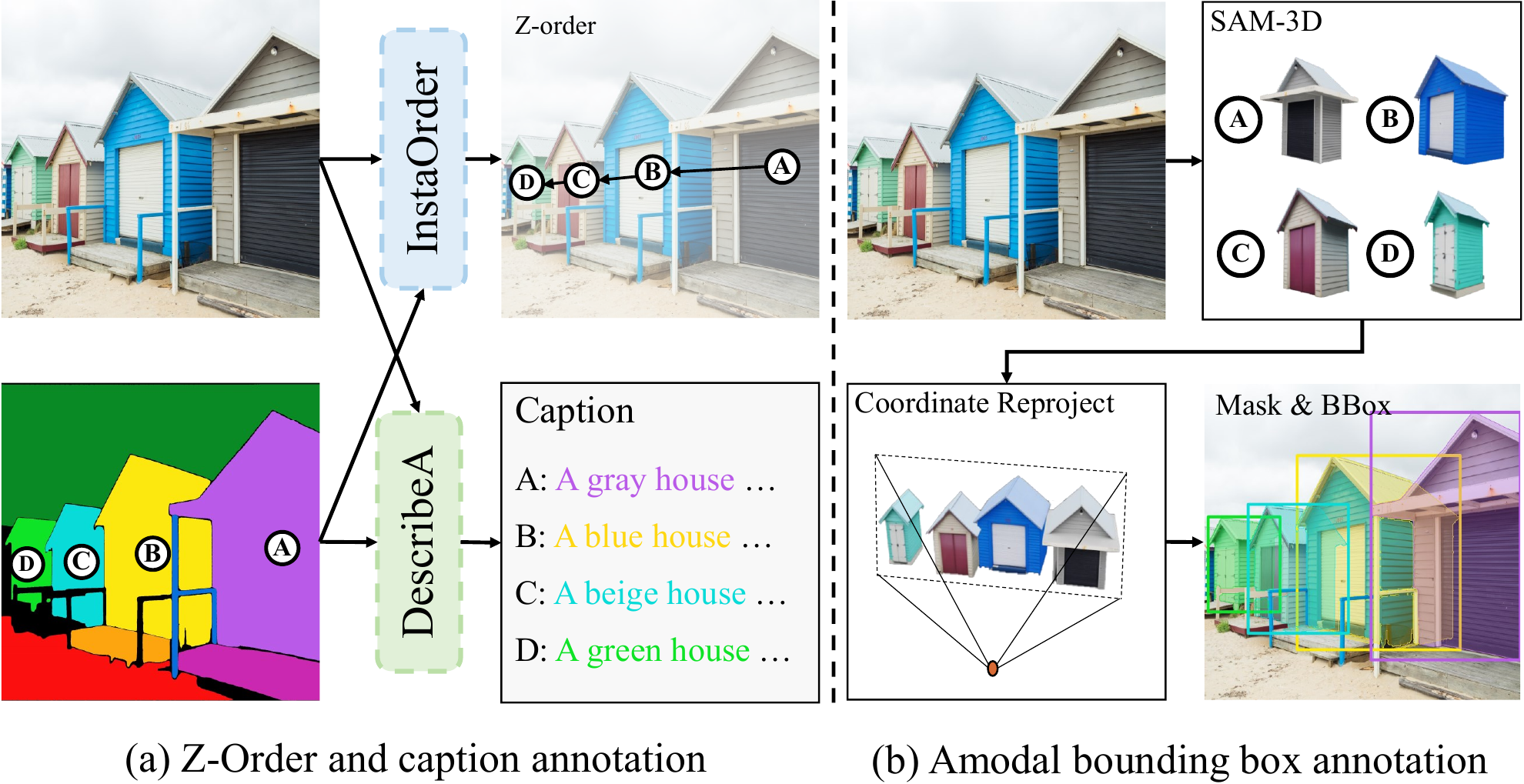}
\vspace{-5mm}
\caption{\textbf{Curation pipeline.} (a) Z-order and captions are annotated via InstaOrder and DescribeAnything. (b) Amodal BBoxes are derived by re-projecting 3D assets reconstructed by SAM-3D.}
\label{fig:dataset_curation}
\vspace{-0.8cm}
\end{figure}

\paragraph{Flow Matching.}
Flow Matching~\cite{lipman2022flow} transports a source distribution $p_0$ (noise) to a target $p_1$ (data). Rectified Flow~\cite{liu2022flow, esser2024scaling} adopts a linear interpolation path $\mathbf{z}_t = t \mathbf{x}_1 + (1-t) \mathbf{x}_0$. The model $v_{\theta}$ predicts the velocity $\mathbf{v}_{target} = \mathbf{x}_1 - \mathbf{x}_0$ by minimizing the following objective:
\vspace{-0.05cm}
\begin{equation}
    \mathcal{L}_{\text{flow}} = \mathbb{E}_{t, \mathbf{x}_0, \mathbf{x}_1} \left[ \| v_{\theta}(\mathbf{z}_t, t) - (\mathbf{x}_1 - \mathbf{x}_0) \|_2^2 \right].
\end{equation}

\vspace{-0.2cm}

\begin{figure*}
    \centering
    \includegraphics[width=0.98\linewidth]{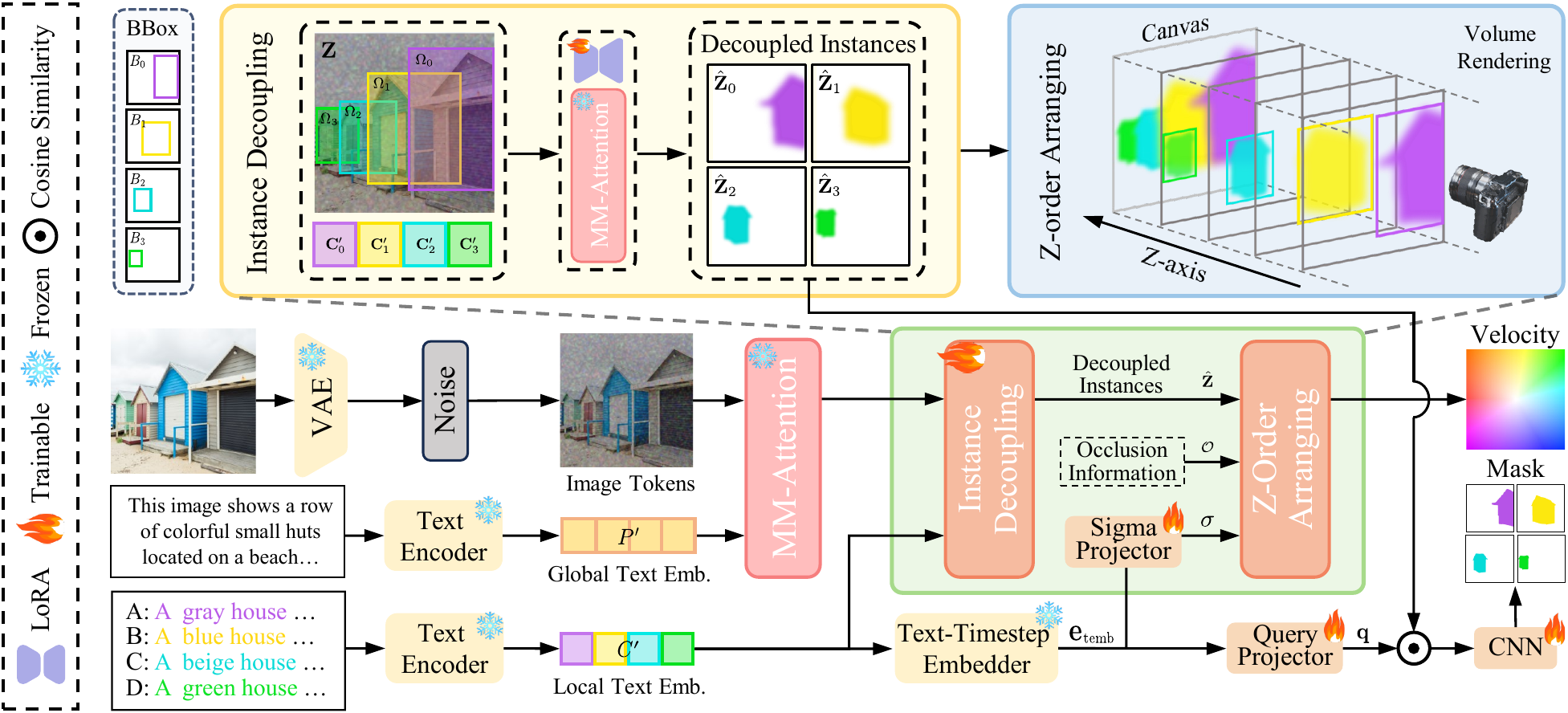}
    \caption{\textbf{The training pipeline of \methodname.} The framework decouples instances and recomposes them using volumetric rendering to resolve occlusions. Simultaneously, a queried alignment mechanism enforce strict spatial consistency via mask supervision.}
    \label{fig:pipeline}
    \vspace{-0.4cm}
\end{figure*}

\subsection{Dataset Curation}
As shown in \cref{fig:dataset_curation}, which illustrates the process with four instances for clarity, our curation pipeline augments the 2D masks of SACap-1M with three critical annotations to support occlusion-aware generation. First, to ensure semantic precision, we employ the pixel-level captioner DescribeAnything~\cite{lian2025describe} to generate instance-specific descriptions strictly based on the mask area, avoiding visual noise from irrelevant adjacent instances. Second, to resolve occlusion ambiguity, we utilize InstaOrder~\cite{lee2022instance} to predict pairwise occlusion relationships, thereby establishing explicit Z-order information. Finally, to recover the full extent of occluded objects for facilitating occlusion supervision, we leverage SAM-3D~\cite{chen2025sam} to lift instances into 3D space. By reconstructing the complete geometry and re-projecting it back to the image plane, we derive amodal mask and bounding boxes. As detailed in \cref{tab:dataset_comparison}, \datasetname\ scales to 1M high-resolution images with 5.7M instances, uniquely featuring open-vocabulary amodal annotations. By incorporating the existing global prompt $P$ from SACap-1M, we define each condition as a quintuple $(M_{i}, B_{i}, \mathcal{O}_{i}, C_{i}, P)$, representing the \textbf{M}ask, \textbf{B}ounding box, \textbf{O}ccluders, instance \textbf{C}aption, and global \textbf{P}rompt.

\subsection{\methodname}
\paragraph{Extending Z-axis via Instance Decoupling.}
Previous methods like Eligen~\cite{zhang2025eligen} and Creatilayout~\cite{zhang2025creatilayout} control instance locations by injecting spatial information directly into the global Multi-Modal Attention (MM-Attention)~\cite{esser2024scaling}. 
However, applying global attention across the entire 2D plane makes it difficult to explicitly model the order information across Z-axis, as all instances and background tokens interact indiscriminately.
To address this, we propose extending the control into the Z-axis by decoupling instances into independent layers.
As shown in \cref{fig:pipeline}, our framework operates in a serial manner. 
Specifically, we derive the visual features $\mathbf{Z} \in \mathbb{R}^{L \times D}$ by processing the image tokens and the computed global prompt embedding $P'$ through the preceding frozen MM-Attention block, where $L$ is the sequence length and $D$ denotes dimension.
For each instance $i$, defined by its bounding box area $B_i$ and caption $C_i$, we identify the subset of token indices $\Omega_i$ that fall within the region $B_i$:
\begin{equation}
    \Omega_i = \{u \mid \text{Coord}(u) \in B_i\},
\end{equation}
where $\text{Coord}(u)$ maps a token index to its 2D spatial coordinates.
Instead of attending to the global context, we extract the local visual sequence $\mathbf{Z}_{\Omega_i} \in \mathbb{R}^{|\Omega_i| \times D}$ corresponding to these indices. 
We then perform MM-Attention strictly between this local visual subset and the specific instance text embeddings $\mathbf{C}_i'$ calculated from instance caption $\mathbf{C}_i$:
\begin{equation}
    \hat{\mathbf{Z}}_{\Omega_i}, \hat{\mathbf{C}_i} = \text{MM-Attention}(\mathbf{Z}_{\Omega_i}, \mathbf{C}_i'),
    \label{eq:decoupled_attn}
\end{equation}
where $\text{MM-Attention}(\cdot, \cdot)$ represents the multi-modal attention reused from previous block and $\hat{\mathbf{Z}}_{\Omega_i}, \hat{\mathbf{C}_i}$ denote the updated features. We further assign that $\hat{\mathbf{Z}}_{i}$ equals $\hat{\mathbf{Z}}_{\Omega_i}$ within $\Omega_i$ and padding zero otherwise.
To adapt the pre-trained backbone for instance control without compromising its original capability, we employ LoRA~\cite{hu2022lora}. We freeze the original parameters and only optimize the injected LoRA layers within the attention projections.
By calculating attention solely within the bounding box scope, we ensure that the visual features of instance $i$ are modulated exclusively by its semantic description, effectively decoupling the generation of different instances before composing them.

\vspace{-0.3cm}

\begin{figure*}[t]
    \centering
    \includegraphics[width=0.98\linewidth]{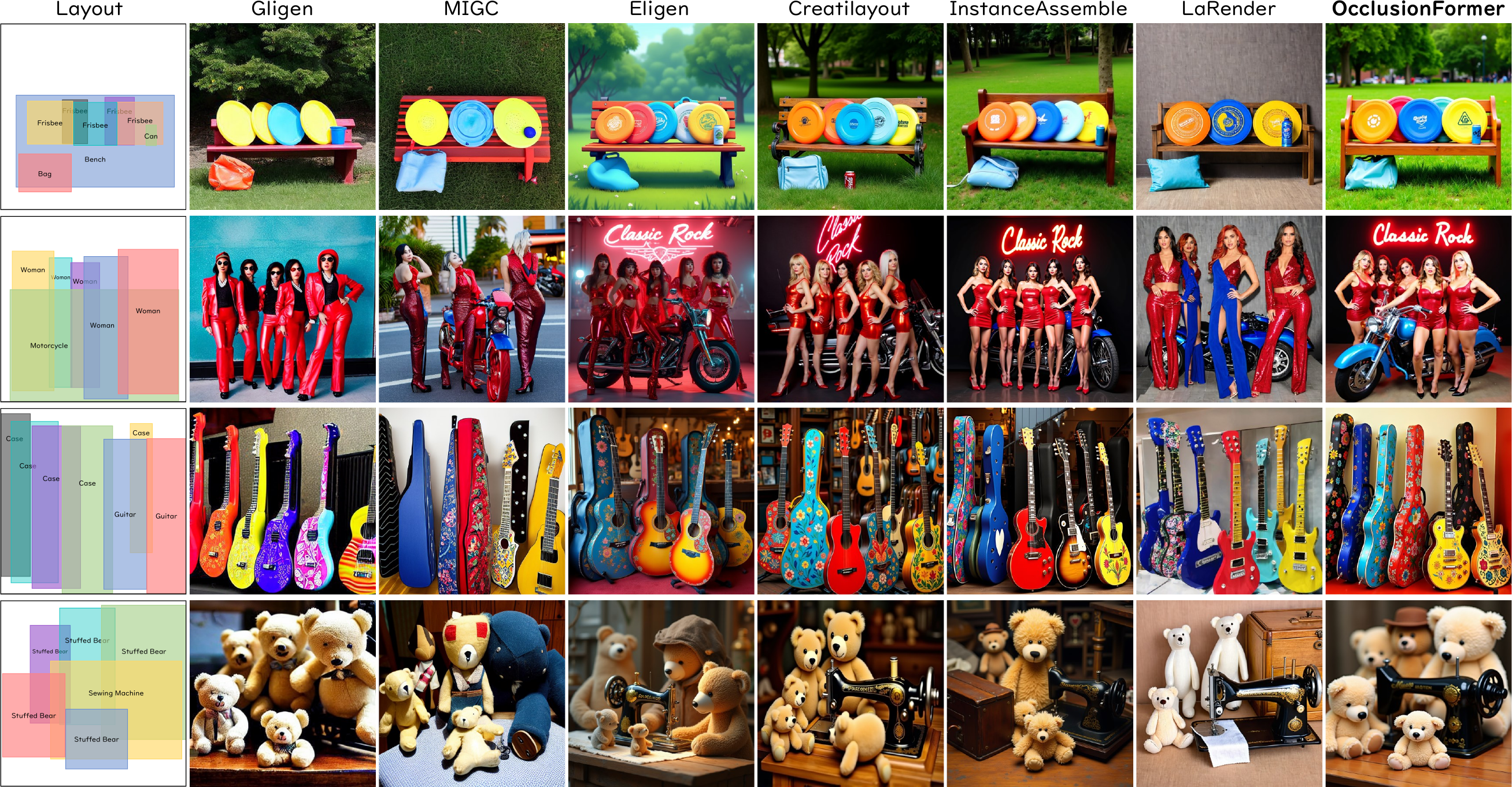}
    \caption{The visual comparison of different methods on the OverLayBench~\cite{li2025overlaybench}.}
    \vspace{-0.4cm}
    \label{fig:qualitative_results}
\end{figure*}

\paragraph{Arranging the Z-order.}
To explicitly model the Z-order, we bring the idea of volume rendering from NeRF~\cite{mildenhall2020nerf}. 
However, to adapt the principle of NeRF for the context of 2D image generation, we follow  LaRender~\cite{zhan2025larender} to view the image plane through a virtual orthogonal camera. 
We conceptualize the composition process as casting rays through the pixel space, arranged according to the provided set of occluder $\mathcal{O}$.
\vspace{-0.15cm}

Drawing inspiration from the modulation vectors in Multimodal Diffusion Transformers~\cite{esser2024scaling}, we predict a learnable vector density $\sigma_i \in \mathbb{R}^D$ for each instance $i$, which is dynamically modulated based on the diffusion state for high-dimensional latent. Specifically, we first compute a conditioning embedding $\mathbf{e}_{\text{temb}}^{i}$ for instance $i$ by fusing the diffusion timestep $t$ and pooled textual projections $y_{i}$ from the text embedding $\mathbf{C}_i'$ via a time-text embedding module:
\vspace{-0.35cm}

\begin{equation}
    \mathbf{e}_{\text{temb}}^{i} = \text{TimeTextEmbed}(t, y_{i}).
\end{equation}
We then project this embedding $\mathbf{e}_{\text{temb}}^{i}$ to obtain $\sigma_i$, effectively allowing the model to adaptively adjust the instance's solidity according to different generation stages.

We then define the opacity $\alpha_i \in \mathbb{R}^D$ at pixel location $\mathbf{p}$ as:
\begin{equation}
    \alpha_i(\mathbf{p}) = \left(1 - \exp(-\sigma_i)\right) \cdot \mathbb{I}(\mathbf{p} \in B_i),
\end{equation}
where $\mathbb{I}(\mathbf{p} \in B_i)$ acts as a binary spatial mask that restricts the instance's opacity to be active only within its bounding box $B_i$.
To handle occlusion, we calculate the transmittance $T_i \in \mathbb{R}^D$, which denotes the probability of light reaching instance $i$ without being blocked. 
Let $\mathcal{O}_i$ be the set of occluders explicitly ordered in front of instance $i$. The transmittance is computed by element-wise operation as:
\vspace{-0.2cm}
\begin{equation}
    T_i(\mathbf{p}) = \exp\left( - \sum_{j \in \mathcal{O}_i} \sigma_j \cdot \mathbb{I}(\mathbf{p} \in B_j) \right).
\end{equation}
This formulation ensures that if a dense occluder $j$ covers the pixel, the transmittance $T_i$ for the background object drops, effectively occluding the background object.
\vspace{-0.1cm}

Finally, we define the rendering weight for instance $i$ as $w_i(\mathbf{p}) = T_i(\mathbf{p}) \cdot \alpha_i(\mathbf{p})$. 
To ensure numerical stability and handle overlaps where no explicit occlusion relationship is defined between instances, we employ a hybrid aggregation strategy. 
For regions with valid occlusion weights, we perform a normalized weighted average. Otherwise, for overlapping regions without occlusion constraints (where only the boxes intersect but objects are non-overlapping), we default to a simple averaging of all features.
The composed feature map $\mathbf{Z}_{out}(\mathbf{p})$ is computed as follows:
\vspace{-0.05cm}
\begin{equation}\vspace{-0.05cm}
    \mathbf{Z}_{out}(\mathbf{p}) = 
    \begin{cases} 
    \frac{\sum_{i} w_i(\mathbf{p}) \cdot \hat{\mathbf{Z}}_{i}(\mathbf{p})}{\sum_{i} w_i(\mathbf{p}) + \epsilon}, & \text{if } \sum_{i} w_i(\mathbf{p}) > 0 \\
    \frac{1}{\max(1, |\mathcal{S}_{\mathbf{p}}|)} \sum_{i \in \mathcal{S}_{\mathbf{p}}} \hat{\mathbf{Z}}_{i}(\mathbf{p}), & \text{otherwise}
    \end{cases}
\end{equation}
where $\mathcal{S}_{\mathbf{p}}$ is the set of bounding boxes of instances covering pixel $\mathbf{p}$, and $\epsilon$ is a small constant for stability. 
Finally, the input feature $\mathbf{Z}$ is added to $\mathbf{Z}_{out}$ via a residual connection.
\vspace{-0.05cm}

\begin{figure*}[t]
    \centering
    \includegraphics[width=0.98\linewidth]{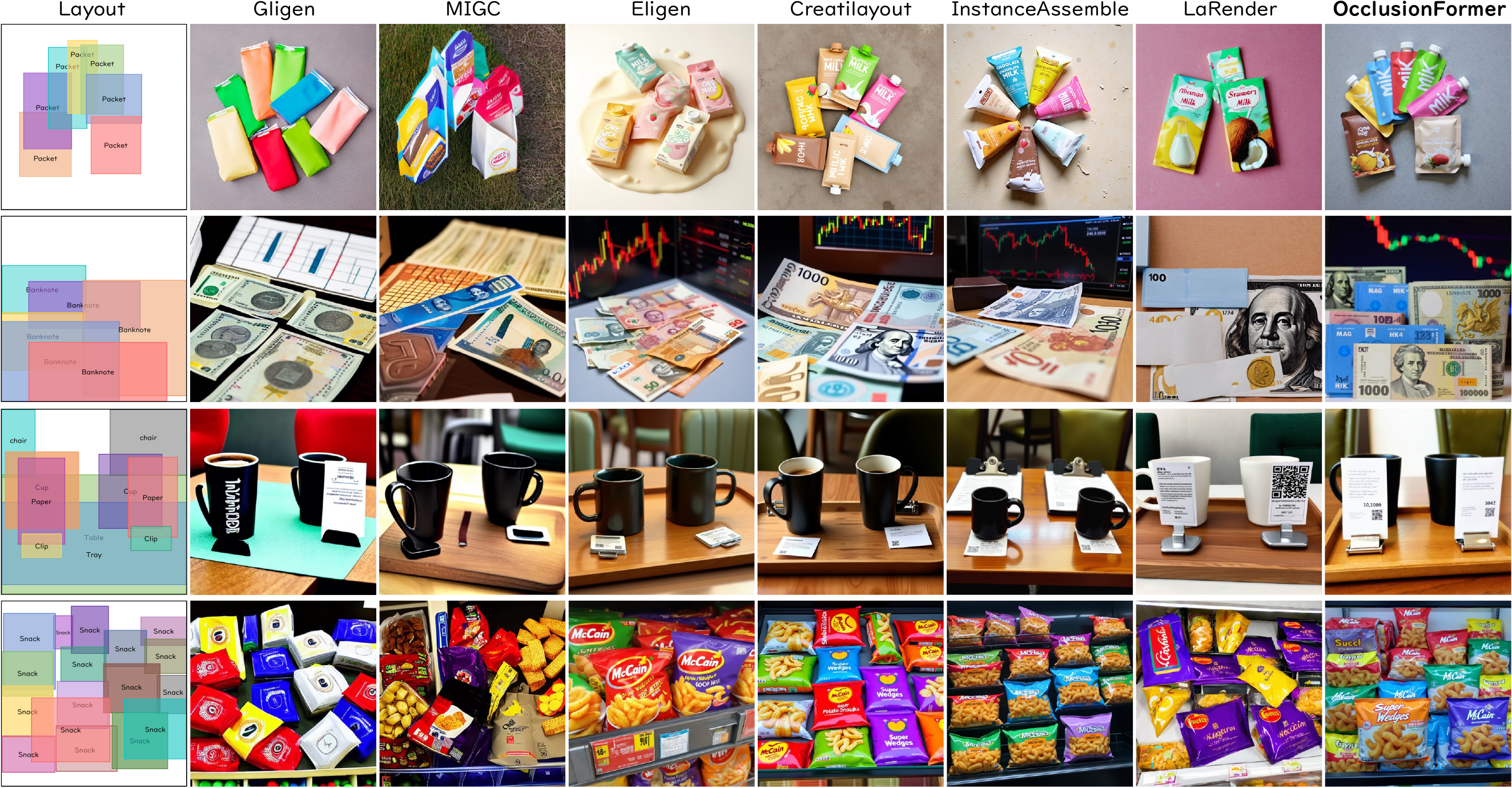}
    \caption{The visual comparison of different methods on our constructed \datasetname Eval.}
    \vspace{-0.4cm}
    \label{fig:qualitative_results_myoverlaybench}
\end{figure*}

\paragraph{Enhancing Alignment via Queried Loss.}
While volumetric rendering resolves occlusion ordering, it relies on the premise that features form coherent geometric structures. To prevent spatial drift and enforce fine-grained shape consistency, we introduce a Queried Alignment Mechanism to explicitly supervise the spatial distribution of features.

For each instance $i$, we derive a learnable query vector $\mathbf{q}_i \in \mathbb{R}^D$ from the time-dependent embedding $\mathbf{e}_{\text{temb}}^{i}$. This query serves as a dynamic semantic anchor, intended to retrieve the spatial footprint of instance from the local visual features $\hat{\mathbf{Z}}_{\Omega_i}$ within $\hat{\mathbf{Z}}_i$. We first compute a spatial similarity map $\mathbf{S}_i \in \mathbb{R}^{H \times W}$ via pixel-wise cosine similarity:
\vspace{-0.1cm}
\begin{equation}
    \mathbf{S}_i(\mathbf{p}) = \frac{\hat{\mathbf{Z}}_{i}(\mathbf{p}) \cdot \mathbf{q}_i}{(\|\hat{\mathbf{Z}}_{i}(\mathbf{p})\| + \epsilon) \|\mathbf{q}_i\|},
\end{equation}
where $\epsilon$ is a small constant. To refine this coarse similarity into a precise shape, we feed $\mathbf{S}_i$ into a lightweight CNN mask predictor $\mathcal{F}_{\theta}$. The predictor outputs a probability map corresponding to background and foreground likelihoods:
\begin{equation}
    \hat{\mathbf{M}}_i = \text{Softmax}\left(\mathcal{F}_{\theta}(\mathbf{S}_i)\right) \in [0, 1]^{H \times W \times 2}.
\end{equation}

\begin{figure}[t]
    \vspace{0.12cm}
    \centering
    \includegraphics[width=\linewidth]{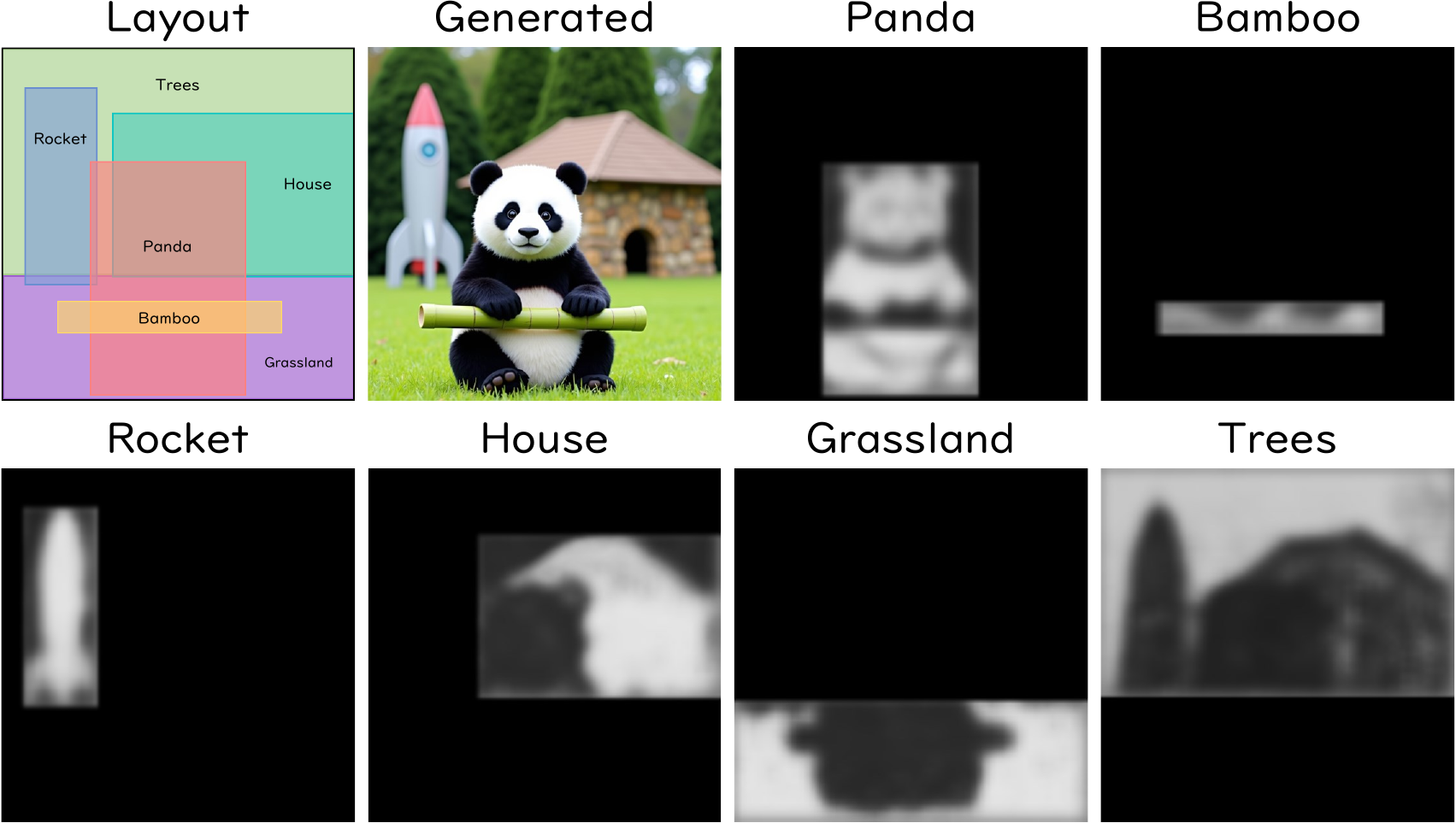}
    \vspace{-0.3cm}
    \caption{Visualization of the predicted foreground probability.}
    \label{fig:mask_pred}
    \vspace{-0.6cm}
\end{figure}

During training, we leverage masks $M_{i}$ provided in \datasetname to enforce alignment via a Cross-Entropy loss $\mathcal{L}_{align}$, which encourages visual features to focus on valid object regions:
\begin{equation}
    \mathcal{L}_{align} = - \frac{1}{N} \sum_{i, \mathbf{p}} \left[ M_{i} \log (\hat{\mathbf{M}}_i^{fg}) + (1 - M_{i}) \log (\hat{\mathbf{M}}_i^{bg}) \right].
\end{equation}
Optimizing this queried loss forces the model to generate features $\hat{\mathbf{Z}}_{\Omega_i}$ that are not only semantically consistent but also aligned with the spatial geometry. As shown in \cref{fig:mask_pred}, the predicted foreground map $\hat{\mathbf{M}}_i^{fg}$ effectively captures the target geometry, validating the efficacy of our supervision.

\begin{table*}[t]
\centering
\small 
\setlength{\tabcolsep}{2pt} 
\caption{Comparison results on Simple, Regular, and Complex subsets on OverLayBench~\cite{li2025overlaybench} and \datasetname Eval.}
\label{tab:results}

\begin{tabularx}{0.98\textwidth}{c|c| *{9}{Y} } 
\toprule
Subset & Method & mIoU$\uparrow$ & O-mIoU$\uparrow$ & SR$_\text{E}$$\uparrow$ & SR$_\text{R}$$\uparrow$ & CLIP-G$\uparrow$ & CLIP-L$\uparrow$ & FID $\downarrow$ & Occ.$\uparrow$ & Dep.$\downarrow$ \\
\midrule

\multirow{7}{*}{\makecell{OverLay-\\Simple}} 
 & GLIGEN           & 0.6380 & 0.3847 & 0.4885 & 0.7849 & 0.3243 & 0.2473 & 36.732 & 0.6055 & 0.2414 \\
 & MIGC             & 0.6009 & 0.3350 & 0.6340 & 0.8044 & 0.3260 & 0.2683 & 33.382 & 0.5631 & 0.2607 \\
 & LaRender         & 0.6604 & 0.4136 & 0.5665 & 0.7767 & 0.2882 & 0.2608 & 38.674 & 0.6294 & 0.2378 \\
 & Eligen           & 0.6673 & 0.4151 & 0.8813 & \underline{0.9165} & 0.3654 & 0.2865 & 27.908 & 0.6823 & 0.2118 \\
 & Creatilayout     & 0.6998 & 0.4725 & 0.8255 & 0.9094 & \textbf{0.3737} & 0.2827 & 25.026 & 0.7559 & 0.1792 \\
 & InstanceAssemble & \underline{0.7279} & \underline{0.5152} & \underline{0.9043} & 0.9105 & 0.3664 & \underline{0.2882} & \underline{24.768} & \underline{0.7852} & \underline{0.1621} \\
 & \textbf{\methodname}                 & \textbf{0.7405} & \textbf{0.5456} & \textbf{0.9241} & \textbf{0.9257} & \underline{0.3711} & \textbf{0.2896} & \textbf{24.596} & \textbf{0.8051} & \textbf{0.1559} \\
\midrule

\multirow{7}{*}{\makecell{OverLay-\\Regular}} 
 & GLIGEN           & 0.5549 & 0.2960 & 0.4577 & 0.7701 & 0.3232 & 0.2346 & 58.122 & 0.5831 & 0.2431 \\
 & MIGC             & 0.4836 & 0.2069 & 0.5663 & 0.7752 & 0.3208 & 0.2530 & 57.290 & 0.4569 & 0.2660 \\
 & LaRender         & 0.5721 & 0.3006 & 0.5497 & 0.7540 & 0.2867 & 0.2607 & 60.935 & 0.5862 & 0.2305 \\
 & Eligen           & 0.5680 & 0.3075 & 0.8437 & 0.8727 & 0.3624 & 0.2712 & 44.839 & 0.6186 & 0.2076 \\
 & Creatilayout     & 0.5997 & 0.3517 & 0.7523 & 0.8604 & \underline{0.3633} & 0.2648 & 43.368 & 0.7124 & 0.1765 \\
 & InstanceAssemble & \underline{0.6299} & \underline{0.3861} & \underline{0.8795} & \underline{0.8746} & 0.3625 & \underline{0.2725} & \underline{43.068} & \underline{0.7475} & \underline{0.1659} \\
 & \textbf{\methodname}                 & \textbf{0.6487} & \textbf{0.4161} & \textbf{0.8822} & \textbf{0.8821} & \textbf{0.3639} & \textbf{0.2745} & \textbf{42.712} & \textbf{0.7811} & \textbf{0.1575} \\
\midrule

\multirow{7}{*}{\makecell{OverLay-\\Complex}} 
 & GLIGEN           & 0.5468 & 0.2763 & 0.4018 & 0.8046 & 0.3219 & 0.2290 & 62.647 & 0.5951 & 0.2251 \\
 & MIGC             & 0.4024 & 0.1367 & 0.4863 & 0.7487 & 0.3132 & 0.2470 & 69.397 & 0.4091 & 0.2968 \\
 & LaRender         & 0.5227 & 0.2507 & 0.4508 & 0.7462 & 0.2685 & 0.2473 & 67.884 & 0.6026 & 0.2374 \\
 & Eligen           & 0.5195 & 0.2569 & 0.7988 & \underline{0.8794} & 0.3604 & 0.2582 & 49.421 & 0.5994 & 0.2378 \\
 & Creatilayout     & 0.5584 & 0.3006 & 0.6923 & 0.8750 & \underline{0.3622} & 0.2532 & 47.793 & \underline{0.7142} & 0.1907 \\
 & InstanceAssemble & \underline{0.5706} & \underline{0.3189} & \underline{0.8348} & 0.8761 & 0.3608 & \textbf{0.2658} & \underline{46.673} & 0.6987 & \underline{0.1791} \\
 & \textbf{\methodname}             & \textbf{0.6037} & \textbf{0.3468} & \textbf{0.8531} & \textbf{0.8890} & \textbf{0.3648} & \underline{0.2640} & \textbf{46.166} & \textbf{0.7797} & \textbf{0.1602} \\
\midrule

\multirow{7}{*}{\makecell{\datasetname\\Eval}} 
 & GLIGEN           & 0.3837 & 0.1695 & 0.7180 & 0.7437 & 0.3074 & 0.2093 & 75.134 & 0.6778 & 0.1805 \\
 & MIGC             & 0.3076 & 0.0958 & 0.6375 & 0.7313 & 0.3003 & 0.2270 & 73.443 & 0.6191 & 0.2531 \\
 & LaRender         & 0.4053 & 0.1709 & 0.7128 & 0.7449 & 0.2673 & 0.2293 & 77.983 & 0.6833 & 0.1790 \\
 & Eligen           & 0.3007 & 0.1016 & 0.8056 & 0.8525 & 0.3487 & 0.2385 & 69.910 & 0.6095 & 0.2533 \\
 & Creatilayout     & 0.4216 & 0.1904 & \underline{0.8129} & \textbf{0.8575} & 0.3501 & \underline{0.2446} & 64.659 & 0.6921 & 0.1837 \\
 & InstanceAssemble & \underline{0.4292} & \underline{0.2021} & 0.8074 & 0.8421 & \underline{0.3512} & 0.2439 & \underline{63.654} & \underline{0.6947} & \underline{0.1711} \\
 & \textbf{\methodname}             & \textbf{0.4509} & \textbf{0.2231} & \textbf{0.8158} & \underline{0.8527} & \textbf{0.3514} & \textbf{0.2466} & \textbf{62.786} & \textbf{0.7568} & \textbf{0.1529} \\
\bottomrule
\end{tabularx}
\vspace{-0.3cm}
\end{table*}

\paragraph{Training Objectives.}
The overall optimization objective combines generative capability with spatial alignment control. We train the model via a weighted sum:
\begin{equation}
    \mathcal{L}_{\text{total}} = \mathcal{L}_{\text{flow}} + \lambda \cdot \mathcal{L}_{\text{align}}.
\end{equation}
Here, $\mathcal{L}_{\text{flow}}$ follows the rectified flow matching formulation~\cite{esser2024scaling}. Given the latent state $\mathbf{z}_t$ at timestep $t$ and conditions $\mathbf{c}$, the network $v_{\theta}$ learns to predict the ground-truth velocity $\mathbf{v}_{target}$:
\begin{equation}
    \mathcal{L}_{\text{flow}} = \mathbb{E}_{t, \mathbf{z}_t, \mathbf{c}} \left[ \| v_{\theta}(\mathbf{z}_t, t, \mathbf{c}) - \mathbf{v}_{target} \|_2^2 \right].
\end{equation}
We empirically set the balancing coefficient \(\lambda = 0.5\) to enforce sufficient geometry constraints without compromising the inherent visual quality of the pre-trained backbone.

\section{Experiment}

\subsection{Experiment Settings}
Our method is built upon Flux.1-dev~\cite{blackforestlabs2024flux} and compared against the previous U-Net-based~\cite{li2023gligen,zhou2024migc,zhan2025larender} and Flux-based~\cite{zhang2025eligen,zhang2025creatilayout,xiang2025instanceassemble} baselines. 

For evaluation, we utilize OverLayBench~\cite{li2025overlaybench}, as it specializes in assessing object occlusion and dense overlaps. To enable more detailed occlusion-aware evaluation, we additionally derive occlusion orders using SAM3~\cite{carion2025sam} and InstaOrder~\cite{lee2022instance}.
However, since OverLayBench consists of synthetic images generated by Flux, a domain gap inevitably exists with real-world scenarios. To address this, we curate an additional \textit{\datasetname Eval} with 1,000 images sampled from our \datasetname, specifically selecting cases with high instance counts and complex occlusion patterns to ensure rigorous realistic evaluation. These samples are excluded in training process.
Following the protocols of OverLayBench, we report metrics across three dimensions:
(1) Spatial Precision: We use mIoU for standard layout accuracy and O-mIoU to specifically evaluate intersection fidelity within complex overlapping regions.
(2) Semantic Consistency: We employ VQA-based SR$_\text{E}$ and SR$_\text{R}$ using Qwen2.5-VL-32B~\cite{bai2025qwen2} to verify entity existence and spatial relationship correctness, respectively. We also report Global (CLIP-G) and Local (CLIP-L) scores~\cite{radford2021learning} for text-image alignment.
(3) Image Quality: FID~\cite{heusel2017gans} is included to assess the realism of generated images.
Additionally, based on the derived occlusion annotations, we report occlusion-aware metrics used in InstaOrder: Occ. (Occlusion Order, measured by F1 score) and Dep. (Depth Order, measured by WHDR~\cite{bell2014intrinsic}), which quantifies the disagreement between predicted and ground truth depth layers.
For implementation, we set LoRA rank to 4 and train for 200K steps with a batch size of 16 and a learning rate of $1e^{-4}$.
\vspace{-2.5mm}

\begin{table*}[t]
\centering
\footnotesize 
\setlength{\tabcolsep}{2pt} 
\caption{Ablation study on the Complex subset of OverLayBench~\cite{li2025overlaybench} and our created \datasetname Eval. We analyze the impact of dynamic density, queried alignment losses, and occlusion conditioning, highlighting the contribution of each component.}

\label{tab:ablation}

\begin{tabularx}{0.98\textwidth}{c|c| *{9}{Y} } 
\toprule
Subset & Method & mIoU$\uparrow$ & O-mIoU$\uparrow$ & SR$_\text{E}$$\uparrow$ & SR$_\text{R}$$\uparrow$ & CLIP-G$\uparrow$ & CLIP-L$\uparrow$ & FID $\downarrow$ & Occ.$\uparrow$ & Dep.$\downarrow$ \\
\midrule

\multirow{7}{*}{\makecell{OverLay-\\Complex}} 
 & w/o Learned Sigma      & 0.5911 & 0.3276 & 0.8482 & 0.8781 & \underline{0.3624} & 0.2617 & 46.258 & 0.7530 & 0.1694 \\
 & w/o Queried Loss       & 0.5922 & 0.3319 & 0.8436 & 0.8798 & 0.3613 & 0.2611 & \textbf{46.094} & 0.7659 & 0.1666 \\
 & w Attn. Map Loss       & 0.5753 & 0.3207 & 0.8353 & 0.8773 & 0.3599 & 0.2602 & 46.433 & 0.7510 & 0.1695 \\
 & w/o Amodal Data        & \underline{0.6004} & \underline{0.3411} & 0.8496 & \underline{0.8855} & 0.3617 & 0.2621 & 46.265 & \underline{0.7703} & \underline{0.1644} \\
 & w/o Inst. Decouple     & 0.5177 & 0.2786 & 0.8043 & 0.8768 & 0.3610 & 0.2569 & 47.734 & 0.6109 & 0.2310 \\
 & w/o Occlusion Cond.    & 0.5912 & 0.3294 & \underline{0.8505} & 0.8800 & 0.3611 & \underline{0.2623} & 46.358 & 0.7262 & 0.1739 \\
 & \textbf{\methodname}   & \textbf{0.6037} & \textbf{0.3468} & \textbf{0.8531} & \textbf{0.8890} & \textbf{0.3648} & \textbf{0.2640} & \underline{46.166} & \textbf{0.7797} & \textbf{0.1602} \\
\midrule

\multirow{7}{*}{\makecell{\datasetname\\Eval}} 
 & w/o Learned Sigma      & 0.4407 & 0.2133 & 0.8084 & 0.8460 & \textbf{0.3523} & 0.2445 & 63.233 & 0.7358 & 0.1586 \\
 & w/o Queried Loss       & 0.4459 & \underline{0.2211} & 0.8024 & 0.8432 & 0.3480 & 0.2436 & 63.213 & 0.7444 & 0.1625 \\
 & w Attn. Map Loss       & 0.4250 & 0.2013 & \underline{0.8128} & 0.8376 & 0.3512 & 0.2405 & 64.125 & 0.7359 & 0.1700 \\
 & w/o Amodal Data        & \underline{0.4462} & 0.2191 & 0.8064 & 0.8462 & 0.3492 & 0.2453 & \underline{62.821} & \underline{0.7491} & \underline{0.1556} \\
 & w/o Inst. Decouple     & 0.3409 & 0.1350 & 0.7645 & 0.8201 & 0.3465 & 0.2295 & 66.576 & 0.6393 & 0.2480 \\
 & w/o Occlusion Cond.    & 0.4417 & 0.2151 & 0.8107 & \underline{0.8503} & 0.3486 & \textbf{0.2471} & 63.505 & 0.7188 & 0.1676 \\
 & \textbf{\methodname}   & \textbf{0.4509} & \textbf{0.2231} & \textbf{0.8158} & \textbf{0.8527} & \underline{0.3514} & \underline{0.2466} & \textbf{62.786} & \textbf{0.7568} & \textbf{0.1529} \\

\bottomrule
\end{tabularx}
\vspace{-8pt}
\end{table*}

\subsection{Experiment Results}
We present the quantitative comparisons on the OverLayBench benchmark~\cite{li2025overlaybench} and \datasetname Eval in \cref{tab:results}. The evaluation is conducted across Simple, Regular, Complex subsets and \datasetname Eval to assess model performance under varying degrees of spatial intricacy. 
To derive the occlusion and depth annotations for evaluation, we first utilize SAM3~\cite{carion2025sam} to segment the generated distinct instances. Subsequently, these segmented instances are fed into the InstaOrderNet and InstaDepthNet modules within the InstaOrder framework~\cite{lee2022instance} to predict the occlusion order and depth order, respectively.

\begin{figure}[t]
    \vspace{0.12cm}
    \centering
    \includegraphics[width=1\linewidth]{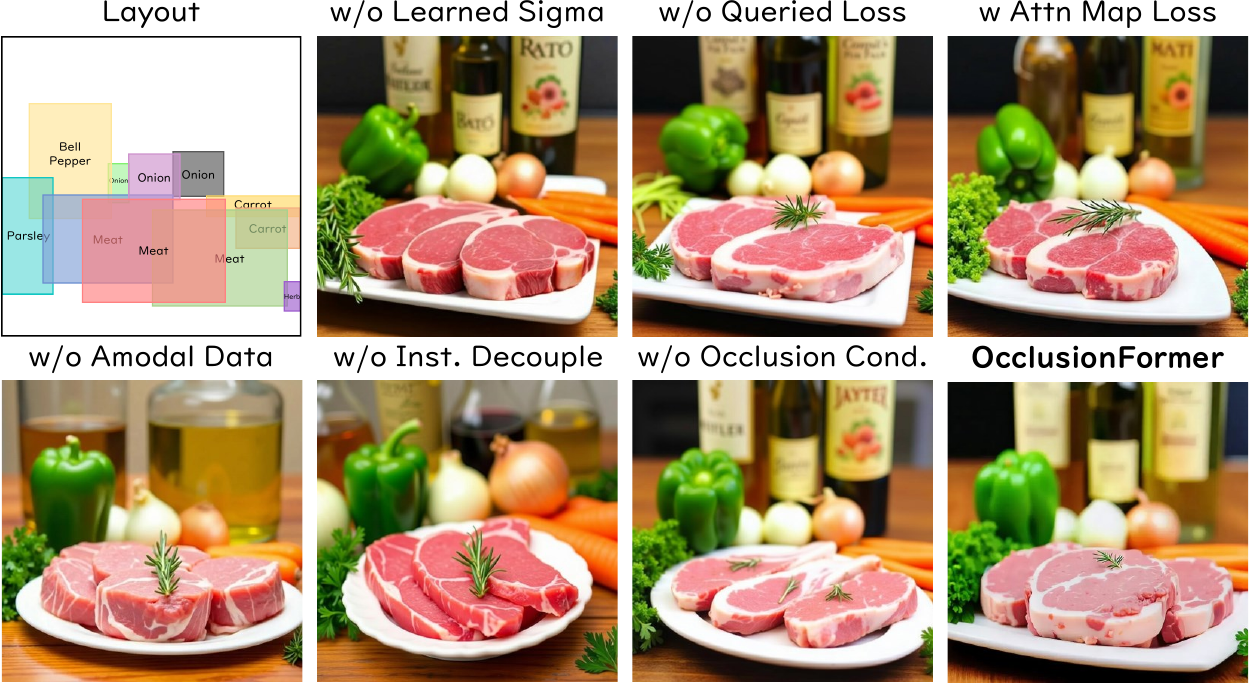}
    \vspace{-0.4cm}
    \caption{Ablation study of different settings of \methodname.}
    \label{fig:ablation}
    \vspace{-0.5cm}
\end{figure}

\noindent\textbf{Qualitative Analysis.}
Visual comparisons on OverLayBench and \datasetname Eval are presented in \cref{fig:qualitative_results} and \cref{fig:qualitative_results_myoverlaybench}, respectively, which reveal that baselines often suffer from object fusion or incorrect Z-order in dense overlap scenes. In contrast, by explicitly modeling Z-axis priority, \methodname\ generates distinct instances with correct occlusion dependencies, maintaining structural integrity.

\noindent\textbf{Z-axis Consistency and Occlusion Handling.}
Our method establishes a new state-of-the-art in occlusion-aware metrics (O-mIoU, Occ., Dep.) across both the OverLayBench and our curated \datasetname Eval. 
This decisive advantage stems from our explicit Z-order modeling via Volumetric Rendering, rather than implicit global attention. By calculating the transmittance $T_i$ derived from the predicted density of occluders, our mechanism effectively suppresses background features in overlapping regions while preserving foreground visibility. This dynamic opacity modulation ensures instances are rendered strictly according to the Z-order, yielding Occ. scores of 0.7797 (Complex) and 0.7568 (\datasetname Eval), demonstrating robustness in challenging scenarios.

\vspace{-1mm}
\noindent\textbf{Spatial Precision and Semantic Alignment.}
Beyond occlusion, our framework excels in 2D layout accuracy and semantic identity, achieving the highest mIoU and O-mIoU scores. We attribute this to the synergy between Instance Decoupling and the Queried Alignment Mechanism. Decoupling the attention computation into local subsets prevents feature bleeding from background tokens. Furthermore, the Queried Alignment loss $\mathcal{L}_{align}$ forces these features to conform to geometry shapes. This filters out noise outside the valid object boundaries, thereby enhancing both the boundary precision and the purity of semantic features (CLIP/SR).

\vspace{-0.2cm}

\begin{figure}[t]
    \vspace{0.12cm}
    \centering
    \includegraphics[width=1\linewidth]{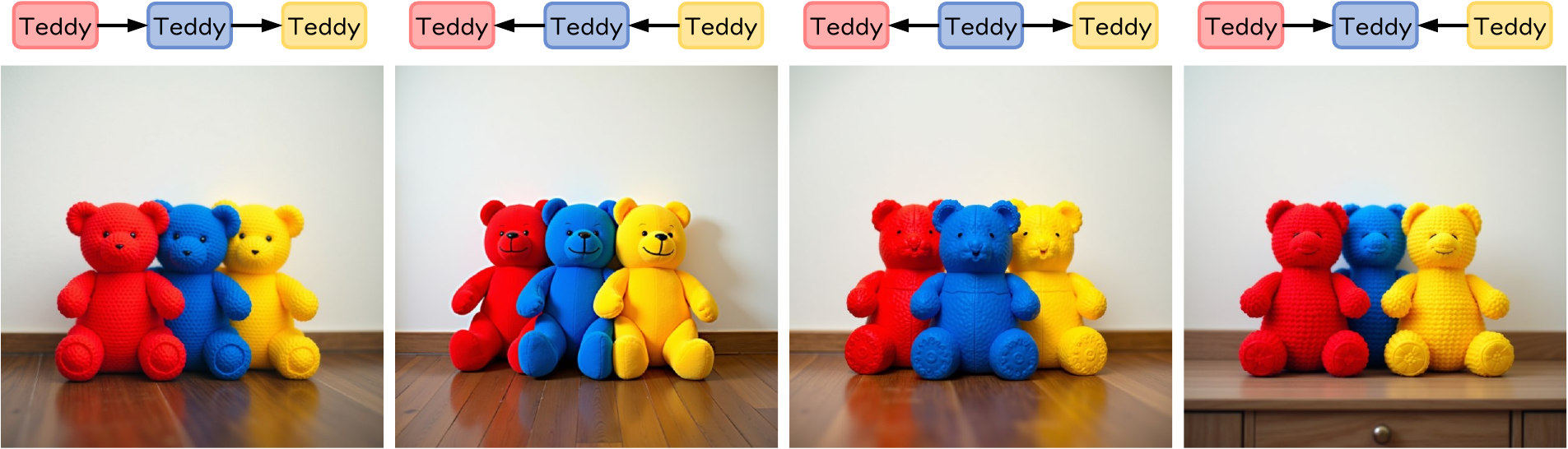}
    \vspace{-0.4cm}
    \caption{Limitations of \methodname. Arrows indicate the direction of occlusion. Best viewed when zoomed in.}
    \label{fig:limitation}
    \vspace{-0.6cm}
\end{figure}

\subsection{Ablation Study}
To validate the effectiveness of \methodname, we conduct ablation studies on both OverLayBench-Complex and \datasetname Eval in \cref{tab:ablation}, and the full results are provided in the appendix. Visual examples are presented in \cref{fig:ablation}.

\noindent\textbf{Significance of Instance Decoupling.}
Reverting to global attention (\textit{w/o Inst. Decouple}) causes the most severe performance collapse across both subsets. The consistent drop in mIoU and Occ. metrics confirms that decoupling is fundamental to prevent feature entanglement between overlapping instances and background tokens, regardless of the domain.

\noindent\textbf{Z-order Modeling and Consistency.}
Removing explicit Z-order (\textit{w/o Occlusion Cond.}) lowers occlusion accuracy on both benchmarks, proving 2D boxes are insufficient for complex overlaps. Similarly, employing a fixed scalar density (specifically, setting $\sigma=5$ in \textit{w/o Learned Sigma}) consistently degrades performance, validating that opacity should be dynamically modulated across diffusion steps to coordinate the transition from noise to structure.

\noindent\textbf{Spatial Alignment and Amodal Data.}
Removing the auxiliary loss (\textit{w/o Queried Loss}) drops O-mIoU, while using a naive BCE loss on the attention map (\textit{w Attn. Map Loss}) further harms performance across datasets, confirming the necessity of our queried loss. Additionally, training without amodal annotations (\textit{w/o Amodal Data}) yields suboptimal results in both settings, indicating that full amodal shapes provide vital geometric signals for learning occlusion.

\vspace{-0.15cm}

\section{Conclusion}
To address the challenge of inter-object occlusion in layout-to-image generation, we introduce \textbf{\datasetname}, a large-scale dataset enriched with explicit Z-order annotations. Building on this, we propose \textbf{\methodname}, an occlusion-aware framework that models the Z-order via volumetric rendering to resolve the order ambiguities in Z-axis and queried alignment to ensure spatial precision. Extensive evaluations on the benchmarks demonstrate that \methodname establishes a new state-of-the-art, significantly outperforming baselines in both occlusion accuracy and visual fidelity.\\
\textbf{Limitations $\&$ Future Work.} 
As illustrated in \cref{fig:limitation}, we generate images using identical layouts and seeds, varying only the occlusion order. While the Z-order arrangement shifts correctly, we observe noticeable inconsistencies in object identity (e.g., variations in the texture and details of the teddy bear). This suggests that the appearance is not fully disentangled from occlusion order. 
Despite this limitation, our method serves as a foundational baseline for occlusion-aware generation, and we envision that future work could further enhance precision and consistency by incorporating post-training strategies, such as Reinforcement Learning.

\section*{Impact Statement}

This paper presents \textit{\methodname}, which improves layout-to-image generation through explicit Z-order control. While the model shares standard risks associated with generative AI, such as inheriting biases from the pre-trained backbone or potential misuse, it significantly enhances structural fidelity in dense overlapped scenes. This capability offers practical benefits for creative design workflows and the generation of high-quality synthetic training data.

\bibliography{ref}

@String(IJCV = {Int. J. Comput. Vis.})

@String(CVPR= {IEEE Conf. Comput. Vis. Pattern Recog.})

@String(ICCV= {Int. Conf. Comput. Vis.})

@String(ECCV= {Eur. Conf. Comput. Vis.})

@String(NIPS= {Adv. Neural Inform. Process. Syst.})

@String(ICML= {Int. Conf. Mach. Learn.})

@String(TOG= {ACM Trans. Graph.})

@String(ICLR = {Int. Conf. Learn. Represent.})

@String(arXiv = {arXiv})

@String(IJCV   = {IJCV})

@String(CVPR  = {CVPR})

@String(ICCV  = {ICCV})

@String(ECCV  = {ECCV})

@String(NIPS  = {NeurIPS})

@String(ICML  = {ICML})

@String(TOG   = {ACM TOG})

@String(ICLR  = {ICLR})

@inproceedings{zhan2025larender,
  title={LaRender: Training-Free Occlusion Control in Image Generation via Latent Rendering},
  author={Zhan, Xiaohang and Liu, Dingming},
  booktitle=ICCV,
  year={2025}
}

@inproceedings{mildenhall2020nerf,
  title={NeRF: Representing Scenes as Neural Radiance Fields for View Synthesis},
  author={Mildenhall, Ben and Srinivasan, Pratul P and Tancik, Matthew and Barron, Jonathan T and Ramamoorthi, Ravi and Ng, Ren},
  booktitle=ECCV,
  year={2020},
}

@article{zhang2025eligen,
  title={Eligen: Entity-level controlled image generation with regional attention},
  author={Zhang, Hong and Duan, Zhongjie and Wang, Xingjun and Chen, Yingda and Zhang, Yu},
  journal=arXiv,
  year={2025}
}

@inproceedings{zhang2025creatilayout,
  title={Creatilayout: Siamese multimodal diffusion transformer for creative layout-to-image generation},
  author={Zhang, Hui and Hong, Dexiang and Wang, Yitong and Shao, Jie and Wu, Xinglong and Wu, Zuxuan and Jiang, Yu-Gang},
  booktitle=ICCV,
  year={2025}
}

@inproceedings{xiang2025instanceassemble,
  title={InstanceAssemble: Layout-Aware Image Generation via Instance Assembling Attention},
  author={Xiang, Qiang and Sun, Shuang and Li, Binglei and Song, Dejia and Li, Huaxia and Chen, Nemo and Tang, Xu and Hu, Yao and Zhang, Junping},
  booktitle=NIPS,
  year={2025}
}

@inproceedings{qin2025scenedesigner,
  title={SceneDesigner: Controllable Multi-Object Image Generation with 9-DoF Pose Manipulation},
  author={Qin, Zhenyuan and Shuai, Xincheng and Ding, Henghui},
  booktitle=NIPS,
  year={2025}
}

@inproceedings{li2023gligen,
  title={Gligen: Open-set grounded text-to-image generation},
  author={Li, Yuheng and Liu, Haotian and Wu, Qingyang and Mu, Fangzhou and Yang, Jianwei and Gao, Jianfeng and Li, Chunyuan and Lee, Yong Jae},
  booktitle=CVPR,
  year={2023}
}

@inproceedings{zhou2024migc,
  title={Migc: Multi-instance generation controller for text-to-image synthesis},
  author={Zhou, Dewei and Li, You and Ma, Fan and Zhang, Xiaoting and Yang, Yi},
  booktitle=CVPR,
  year={2024}
}

@inproceedings{liseg2any,
  title={Seg2Any: Open-set Segmentation-Mask-to-Image Generation with Precise Shape and Semantic Control},
  author={Li, Danfeng and Zhang, Hui and Wang, Sheng and Li, Jiacheng and Wu, Zuxuan},
  booktitle=NIPS,
  year={2025}
}

@inproceedings{lv2024place,
  title={Place: Adaptive layout-semantic fusion for semantic image synthesis},
  author={Lv, Zhengyao and Wei, Yuxiang and Zuo, Wangmeng and Wong, Kwan-Yee K},
  booktitle=CVPR,
  year={2024}
}

@inproceedings{kirillov2023segment,
  title={Segment anything},
  author={Kirillov, Alexander and Mintun, Eric and Ravi, Nikhila and Mao, Hanzi and Rolland, Chloe and Gustafson, Laura and Xiao, Tete and Whitehead, Spencer and Berg, Alexander C and Lo, Wan-Yen and others},
  booktitle=ICCV,
  year={2023}
}

@article{chen2025sam,
  title={SAM 3D: 3Dfy Anything in Images},
  author={Chen, Xingyu and Chu, Fu-Jen and Gleize, Pierre and Liang, Kevin J and Sax, Alexander and Tang, Hao and Wang, Weiyao and Guo, Michelle and Hardin, Thibaut and Li, Xiang and others},
  journal=arXiv,
  year={2025}
}

@inproceedings{esser2024scaling,
  title={Scaling rectified flow transformers for high-resolution image synthesis},
  author={Esser, Patrick and Kulal, Sumith and Blattmann, Andreas and Entezari, Rahim and M{\"u}ller, Jonas and Saini, Harry and Levi, Yam and Lorenz, Dominik and Sauer, Axel and Boesel, Frederic and others},
  booktitle=ICML,
  year={2024}
}

@misc{blackforestlabs2024flux,
  title = {FLUX.1-dev},
  author = {Black Forest Labs},
  year = {2024},
}

@inproceedings{xie2023boxdiff,
  title={Boxdiff: Text-to-image synthesis with training-free box-constrained diffusion},
  author={Xie, Jinheng and Li, Yuexiang and Huang, Yawen and Liu, Haozhe and Zhang, Wentian and Zheng, Yefeng and Shou, Mike Zheng},
  booktitle=ICCV,
  year={2023}
}

@article{bar2023multidiffusion,
  title={Multidiffusion: Fusing diffusion paths for controlled image generation},
  author={Bar-Tal, Omer and Yariv, Lior and Lipman, Yaron and Dekel, Tali},
  journal=arXiv,
  year={2023}
}

@inproceedings{lin2014microsoft,
  title={Microsoft coco: Common objects in context},
  author={Lin, Tsung-Yi and Maire, Michael and Belongie, Serge and Hays, James and Perona, Pietro and Ramanan, Deva and Doll{\'a}r, Piotr and Zitnick, C Lawrence},
  booktitle=ECCV,
  year={2014}
}

@inproceedings{lian2025describe,
  title={Describe anything: Detailed localized image and video captioning},
  author={Lian, Long and Ding, Yifan and Ge, Yunhao and Liu, Sifei and Mao, Hanzi and Li, Boyi and Pavone, Marco and Liu, Ming-Yu and Darrell, Trevor and Yala, Adam and others},
  booktitle=ICCV,
  year={2025}
}

@inproceedings{lee2022instance,
  title={Instance-wise occlusion and depth orders in natural scenes},
  author={Lee, Hyunmin and Park, Jaesik},
  booktitle=CVPR,
  year={2022}
}

@article{kuznetsova2020open,
  title={The open images dataset v4: Unified image classification, object detection, and visual relationship detection at scale},
  author={Kuznetsova, Alina and Rom, Hassan and Alldrin, Neil and Uijlings, Jasper and Krasin, Ivan and Pont-Tuset, Jordi and Kamali, Shahab and Popov, Stefan and Malloci, Matteo and Kolesnikov, Alexander and others},
  journal=IJCV,
  year={2020}
}

@article{krishna2017visual,
  title={Visual genome: Connecting language and vision using crowdsourced dense image annotations},
  author={Krishna, Ranjay and Zhu, Yuke and Groth, Oliver and Johnson, Justin and Hata, Kenji and Kravitz, Joshua and Chen, Stephanie and Kalantidis, Yannis and Li, Li-Jia and Shamma, David A and others},
  journal=IJCV,
  year={2017}
}

@article{lipman2022flow,
  title={Flow matching for generative modeling},
  author={Lipman, Yaron and Chen, Ricky TQ and Ben-Hamu, Heli and Nickel, Maximilian and Le, Matt},
  journal=arXiv,
  year={2022}
}

@inproceedings{hu2022lora,
  title={Lora: Low-rank adaptation of large language models},
  author={Hu, Edward J and Shen, Yelong and Wallis, Phillip and Allen-Zhu, Zeyuan and Li, Yuanzhi and Wang, Shean and Wang, Lu and Chen, Weizhu and others},
  booktitle=ICLR,
  year={2022}
}

@inproceedings{liu2022flow,
  title={Flow straight and fast: Learning to generate and transfer data with rectified flow},
  author={Liu, Xingchao and Gong, Chengyue and Liu, Qiang},
  booktitle=NIPS,
  year={2022}
}

@inproceedings{li2025overlaybench,
  title={OverLayBench: A Benchmark for Layout-to-Image Generation with Dense Overlaps},
  author={Li, Bingnan and Wang, Chen-Yu and Xu, Haiyang and Zhang, Xiang and Armand, Ethan and Srivastava, Divyansh and Shan, Xiaojun and Chen, Zeyuan and Xie, Jianwen and Tu, Zhuowen},
  booktitle=NIPS,
  year={2025}
}

@article{bell2014intrinsic,
  title={Intrinsic images in the wild},
  author={Bell, Sean and Bala, Kavita and Snavely, Noah},
  journal=TOG,
  year={2014}
}

@article{bai2025qwen2,
  title={Qwen2. 5-vl technical report},
  author={Bai, Shuai and Chen, Keqin and Liu, Xuejing and Wang, Jialin and Ge, Wenbin and Song, Sibo and Dang, Kai and Wang, Peng and Wang, Shijie and Tang, Jun and others},
  journal=arXiv,
  year={2025}
}

@article{carion2025sam,
  title={Sam 3: Segment anything with concepts},
  author={Carion, Nicolas and Gustafson, Laura and Hu, Yuan-Ting and Debnath, Shoubhik Friend and Hu, Ronghang and Suris, Didac and Ryali, Chaitanya and Alwala, Kalyan Vasudev and Khedr, Haitham and Huang, Andrew and others},
  journal=arXiv,
  year={2025}
}

@inproceedings{zhu2017semantic,
  title={Semantic amodal segmentation},
  author={Zhu, Yan and Tian, Yuandong and Metaxas, Dimitris and Doll{\'a}r, Piotr},
  booktitle=CVPR,
  year={2017}
}

@inproceedings{li2025control,
  title={Control and Realism: Best of Both Worlds in Layout-to-Image without Training},
  author={Li, Bonan and Hu, Yinhan and Liu, Songhua and Wang, Xinchao},
  booktitle=ICML,
  year={2025}
}

@inproceedings{he2025plangen,
  title={PlanGen: Towards Unified Layout Planning and Image Generation in Auto-Regressive Vision Language Models},
  author={He, Runze and Cheng, Bo and others},
  booktitle=ICCV,
  year={2025}
}

@inproceedings{li2025anyi2v,
  title={AnyI2V: Animating Any Conditional Image with Motion Control},
  author={Li, Ziye and Luo, Hao and Shuai, Xincheng and Ding, Henghui},
  booktitle=ICCV,
  year={2025}
}

@inproceedings{sun2024anycontrol,
  title={Anycontrol: create your artwork with versatile control on text-to-image generation},
  author={Sun, Yanan and Liu, Yanchen and Tang, Yinhao and Pei, Wenjie and Chen, Kai},
  booktitle=ECCV,
  year={2024}
}

@inproceedings{zhang2023adding,
  title={Adding conditional control to text-to-image diffusion models},
  author={Zhang, Lvmin and Rao, Anyi and Agrawala, Maneesh},
  booktitle=ICCV,
  year={2023}
}

@inproceedings{wang2024instancediffusion,
  title={Instancediffusion: Instance-level control for image generation},
  author={Wang, Xudong and Darrell, Trevor and Rambhatla, Sai Saketh and Girdhar, Rohit and Misra, Ishan},
  booktitle=CVPR,
  year={2024}
}

@inproceedings{cheng2024hico,
  title={Hico: Hierarchical controllable diffusion model for layout-to-image generation},
  author={Cheng, Bo and Ma, Yuhang and Wu, Liebucha and Liu, Shanyuan and Ma, Ao and Wu, Xiaoyu and Leng, Dawei and Yin, Yuhui},
  booktitle=NIPS,
  year={2024}
}

@article{chen2024region,
  title={Region-aware text-to-image generation via hard binding and soft refinement},
  author={Chen, Zhennan and Li, Yajie and Wang, Haofan and Chen, Zhibo and Jiang, Zhengkai and Li, Jun and Wang, Qian and Yang, Jian and Tai, Ying},
  journal=arXiv,
  year={2024}
}

@inproceedings{lin2024ctrl,
  title={Ctrl-x: Controlling structure and appearance for text-to-image generation without guidance},
  author={Lin, Kuan Heng and Mo, Sicheng and Klingher, Ben and Mu, Fangzhou and Zhou, Bolei},
  booktitle=NIPS,
  year={2024}
}

@inproceedings{mo2024freecontrol,
  title={Freecontrol: Training-free spatial control of any text-to-image diffusion model with any condition},
  author={Mo, Sicheng and Mu, Fangzhou and Lin, Kuan Heng and Liu, Yanli and Guan, Bochen and Li, Yin and Zhou, Bolei},
  booktitle=CVPR,
  year={2024}
}

@inproceedings{heusel2017gans,
  title={Gans trained by a two time-scale update rule converge to a local nash equilibrium},
  author={Heusel, Martin and Ramsauer, Hubert and Unterthiner, Thomas and Nessler, Bernhard and Hochreiter, Sepp},
  booktitle=NIPS,
  year={2017}
}

@inproceedings{radford2021learning,
  title={Learning transferable visual models from natural language supervision},
  author={Radford, Alec and Kim, Jong Wook and Hallacy, Chris and Ramesh, Aditya and Goh, Gabriel and Agarwal, Sandhini and Sastry, Girish and Askell, Amanda and Mishkin, Pamela and Clark, Jack and others},
  booktitle=ICML,
  year={2021}
}
\bibliographystyle{icml2026}

\clearpage
\newpage
\appendix

\newcommand{\appendixteaser}{
    \centering
    \includegraphics[width=\textwidth]{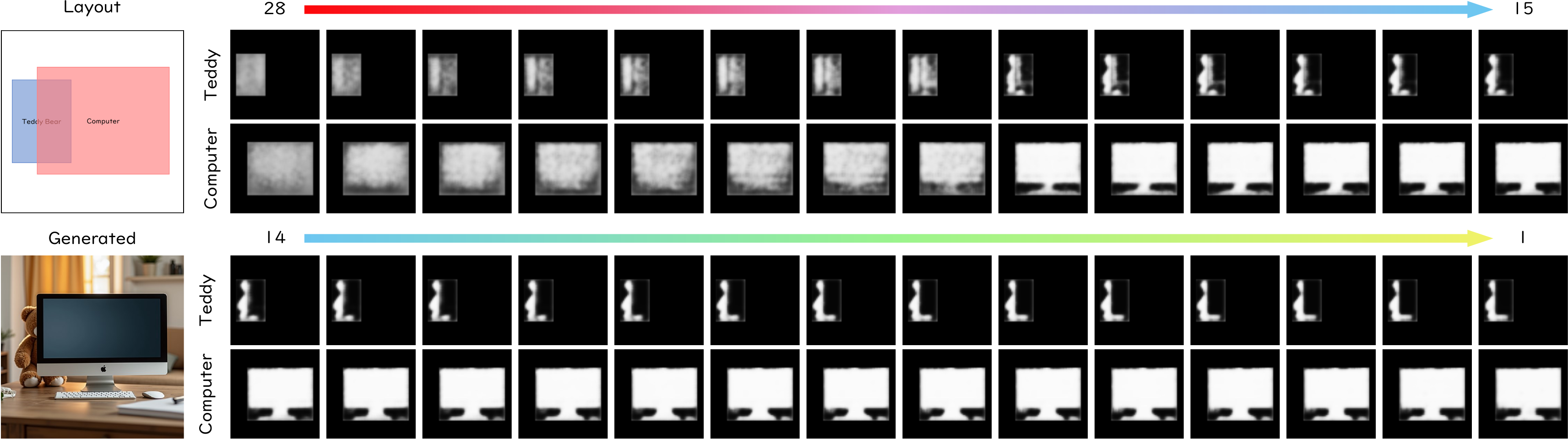}
    \captionof{figure}{Progression of predicted masks during the denoising process, with the total number of timesteps set to 28.}
    \label{fig:supp_mask_pred}
}

\twocolumn[
    \hrule height 1pt
    
    \vspace{0.15in}
    \centering
    {\Large \bf Appendix for: \\ \methodname: Arranging Z-Order for Layout-Grounded Image Generation\par}
    \vspace{0.15in}
    
    \hrule height 1pt
    
    \vspace{0.15in}
    \appendixteaser
    \vspace{0.5cm}
]

\begin{table*}[t]
\centering
\small 
\setlength{\tabcolsep}{2pt} 
\caption{Full ablation study on the Simple, Regular, Complex subsets of OverLayBench~\cite{li2025overlaybench} and our \datasetname Eval.}
\label{tab:ablation_full}

\begin{tabularx}{0.98\textwidth}{c|c| *{9}{Y} } 
\toprule
Subset & Method & mIoU$\uparrow$ & O-mIoU$\uparrow$ & SR$_\text{E}$$\uparrow$ & SR$_\text{R}$$\uparrow$ & CLIP-G$\uparrow$ & CLIP-L$\uparrow$ & FID $\downarrow$ & Occ.$\uparrow$ & Dep.$\downarrow$ \\
\midrule

\multirow{8}{*}{\makecell{OverLay-\\Simple}} 
 & \textcolor{gray}{Flux.1-dev} & \textcolor{gray}{0.3958} & \textcolor{gray}{0.1937} & \textcolor{gray}{0.7878} & \textcolor{gray}{0.8959} & \textcolor{gray}{0.3746} & \textcolor{gray}{0.2475} & \textcolor{gray}{24.282} & \textcolor{gray}{0.5378} & \textcolor{gray}{0.2586} \\
 & w/o Learned Sigma      & 0.7346 & 0.5391 & 0.9182 & \textbf{0.9287} & 0.3673 & \underline{0.2895} & 24.716 & 0.7883 & 0.1588 \\
 & w/o Queried Loss       & \underline{0.7390} & 0.5347 & 0.9130 & \underline{0.9283} & 0.3674 & 0.2889 & 24.763 & 0.7919 & 0.1624 \\
 & w Attn. Map Loss       & 0.7196 & 0.5042 & 0.9096 & 0.9188 & 0.3665 & 0.2875 & 24.880 & 0.7833 & 0.1681 \\
 & w/o Amodal Data        & 0.7368 & \textbf{0.5460} & 0.9234 & 0.9268 & \underline{0.3703} & 0.2880 & \underline{24.630} & \underline{0.8013} & \underline{0.1583} \\
 & w/o Inst. Decouple     & 0.6844 & 0.4879 & 0.8975 & 0.9142 & 0.3682 & 0.2830 & 25.264 & 0.7050 & 0.1954 \\
 & w/o Occlusion Cond.    & 0.7385 & 0.5405 & \underline{0.9235} & 0.9204 & 0.3659 & 0.2885 & 24.796 & 0.7822 & 0.1606 \\
 & \textbf{\methodname}   & \textbf{0.7405} & \underline{0.5456} & \textbf{0.9241} & 0.9257 & \textbf{0.3711} & \textbf{0.2896} & \textbf{24.596} & \textbf{0.8051} & \textbf{0.1559} \\
\midrule

\multirow{8}{*}{\makecell{OverLay-\\Regular}} 
 & \textcolor{gray}{Flux.1-dev} & \textcolor{gray}{0.3250} & \textcolor{gray}{0.1410} & \textcolor{gray}{0.7223} & \textcolor{gray}{0.8434} & \textcolor{gray}{0.3704} & \textcolor{gray}{0.2327} & \textcolor{gray}{43.670} & \textcolor{gray}{0.5081} & \textcolor{gray}{0.2562} \\
 & w/o Learned Sigma      & 0.6321 & 0.4046 & 0.8767 & \underline{0.8819} & 0.3626 & 0.2742 & 42.881 & 0.7542 & 0.1591 \\
 & w/o Queried Loss       & 0.6367 & 0.4098 & 0.8714 & 0.8761 & 0.3630 & 0.2738 & 42.779 & 0.7629 & 0.1599 \\
 & w Attn. Map Loss       & 0.6150 & 0.3866 & 0.8689 & 0.8727 & 0.3609 & 0.2724 & 43.198 & 0.7536 & 0.1594 \\
 & w/o Amodal Data        & \underline{0.6454} & \underline{0.4132} & 0.8798 & 0.8803 & \textbf{0.3644} & \underline{0.2744} & \textbf{42.679} & \underline{0.7793} & 0.1623 \\
 & w/o Inst. Decouple     & 0.5843 & 0.3456 & 0.8573 & 0.8718 & 0.3605 & 0.2675 & 45.668 & 0.6728 & 0.2334 \\
 & w/o Occlusion Cond.    & 0.6301 & 0.4028 & \textbf{0.8824} & 0.8801 & 0.3616 & 0.2750 & 43.412 & 0.7691 & \underline{0.1586} \\
 & \textbf{\methodname}   & \textbf{0.6487} & \textbf{0.4161} & \underline{0.8822} & \textbf{0.8821} & \underline{0.3639} & \textbf{0.2745} & \underline{42.712} & \textbf{0.7811} & \textbf{0.1575} \\
\midrule

\multirow{8}{*}{\makecell{OverLay-\\Complex}} 
 & \textcolor{gray}{Flux.1-dev} & \textcolor{gray}{0.3342} & \textcolor{gray}{0.1402} & \textcolor{gray}{0.6345} & \textcolor{gray}{0.8695} & \textcolor{gray}{0.3706} & \textcolor{gray}{0.2276} & \textcolor{gray}{46.609} & \textcolor{gray}{0.4611} & \textcolor{gray}{0.2846} \\
 & w/o Learned Sigma      & 0.5911 & 0.3276 & 0.8482 & 0.8781 & \underline{0.3624} & 0.2617 & 46.258 & 0.7530 & 0.1694 \\
 & w/o Queried Loss       & 0.5922 & 0.3319 & 0.8436 & 0.8798 & 0.3613 & 0.2611 & \textbf{46.094} & 0.7659 & 0.1666 \\
 & w Attn. Map Loss       & 0.5753 & 0.3207 & 0.8353 & 0.8773 & 0.3599 & 0.2602 & 46.433 & 0.7510 & 0.1695 \\
 & w/o Amodal Data        & \underline{0.6004} & \underline{0.3411} & 0.8496 & \underline{0.8855} & 0.3617 & 0.2621 & 46.265 & \underline{0.7703} & \underline{0.1644} \\
 & w/o Inst. Decouple     & 0.5177 & 0.2786 & 0.8043 & 0.8768 & 0.3610 & 0.2569 & 47.734 & 0.6109 & 0.2310 \\
 & w/o Occlusion Cond.    & 0.5912 & 0.3294 & \underline{0.8505} & 0.8800 & 0.3611 & \underline{0.2623} & 46.358 & 0.7262 & 0.1739 \\
 & \textbf{\methodname}   & \textbf{0.6037} & \textbf{0.3468} & \textbf{0.8531} & \textbf{0.8890} & \textbf{0.3648} & \textbf{0.2640} & \underline{46.166} & \textbf{0.7797} & \textbf{0.1602} \\
\midrule

\multirow{8}{*}{\makecell{\datasetname\\Eval}} 
 & \textcolor{gray}{Flux.1-dev} & \textcolor{gray}{0.1887} & \textcolor{gray}{0.0536} & \textcolor{gray}{0.7845} & \textcolor{gray}{0.8202} & \textcolor{gray}{0.3525} & \textcolor{gray}{0.2127} & \textcolor{gray}{70.226} & \textcolor{gray}{0.4269} & \textcolor{gray}{0.2944} \\
 & w/o Learned Sigma      & 0.4407 & 0.2133 & 0.8084 & 0.8460 & \textbf{0.3523} & 0.2445 & 63.233 & 0.7358 & 0.1586 \\
 & w/o Queried Loss       & 0.4459 & \underline{0.2211} & 0.8024 & 0.8432 & 0.3480 & 0.2436 & 63.213 & 0.7444 & 0.1625 \\
 & w Attn. Map Loss       & 0.4250 & 0.2013 & \underline{0.8128} & 0.8376 & 0.3512 & 0.2405 & 64.125 & 0.7359 & 0.1700 \\
 & w/o Amodal Data        & \underline{0.4462} & 0.2191 & 0.8064 & 0.8462 & 0.3492 & 0.2453 & \underline{62.821} & \underline{0.7491} & \underline{0.1556} \\
 & w/o Inst. Decouple     & 0.3409 & 0.1350 & 0.7645 & 0.8201 & 0.3465 & 0.2295 & 66.576 & 0.6393 & 0.2480 \\
 & w/o Occlusion Cond.    & 0.4417 & 0.2151 & 0.8107 & \underline{0.8503} & 0.3486 & \textbf{0.2471} & 63.505 & 0.7188 & 0.1676 \\
 & \textbf{\methodname}   & \textbf{0.4509} & \textbf{0.2231} & \textbf{0.8158} & \textbf{0.8527} & \underline{0.3514} & \underline{0.2466} & \textbf{62.786} & \textbf{0.7568} & \textbf{0.1529} \\

\bottomrule
\end{tabularx}
\vspace{-8pt}
\end{table*}

\section{More Implementation Details}

\paragraph{Conditioning Projections and Softplus Activation.}
To derive instance-specific control parameters, we employ an adaptive projection module. This module processes the time-dependent text embedding through a SiLU activation followed by two parallel Linear layers. One Linear layer projects the semantic query vector $\mathbf{q}_i$ to retrieve spatial alignment features via cosine similarity. The other Linear layer predicts the raw density value for the instance. To strictly enforce the physical constraint that optical density must be non-negative, we apply the Softplus activation function to the raw output of the density projection layer:
\vspace{-2mm}

\begin{equation}
    \sigma_i = \text{Softplus}(\text{Linear}(\text{SiLU}(\mathbf{e}_{\text{temb}}))).
\end{equation}

\paragraph{Mask Predictor Architecture.}
The mask predictor, designed to refine the coarse spatial similarity map into precise foreground-background probability, is implemented as a lightweight Convolutional Neural Network (CNN). It takes a single-channel similarity map as input and processes it through the following structure:
\begin{enumerate}
    \item A $3\times3$ convolution ($1 \to 32$ channels, padding 1) followed by GELU activation;
    \item A $3\times3$ convolution ($32 \to 16$ channels, padding 1) followed by GELU activation;
    \item A $1\times1$ convolution ($16 \to 2$ channels) to output the logits for background and foreground probabilities.
\end{enumerate}

\paragraph{Training and Inference Strategy.}
During training, we adopt a time-dependent mask supervision strategy. Specifically, we utilize amodal masks as the supervision target during high noise levels (e.g., $t \in [700, 1000]$) to establish global structure, and switch to modal (visible) masks for the remaining steps (e.g., $t < 700$). 
This curriculum encourages the model to reconstruct complete amodal features in the early phase to facilitate occlusion learning, while prioritizing precise visible boundary refinement in the later stages.
During inference, we employ a 28-step denoising schedule. Following foundation work~\cite{li2023gligen}, layout guidance is activated exclusively during the initial 30\% of the denoising process for the balance of quality and speed. All our experiments are implemented on Nvidia A800 GPUs.

\vspace{-2.5mm}

\begin{figure*}[t]
    \centering
    \begin{minipage}{\linewidth}
        \centering
        \includegraphics[width=0.98\linewidth]{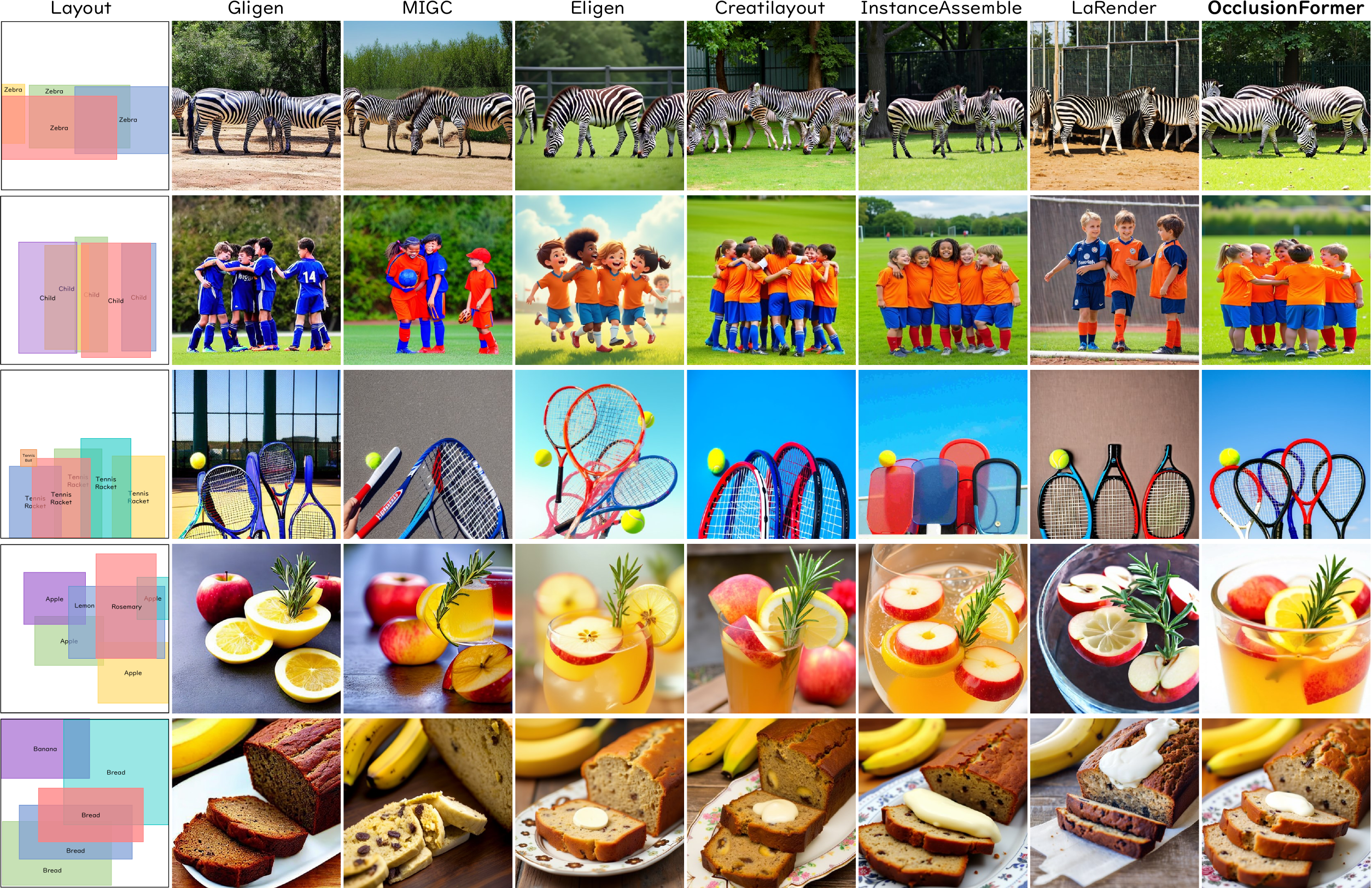}
        \caption{The visual comparison of different methods on the OverLayBench~\cite{li2025overlaybench}.}
        \label{fig:supp_qualitative_results_overlaybench}
    \end{minipage}
    
    \vspace{0.5cm}

    \begin{minipage}{\linewidth}
        \centering
        \includegraphics[width=0.98\linewidth]{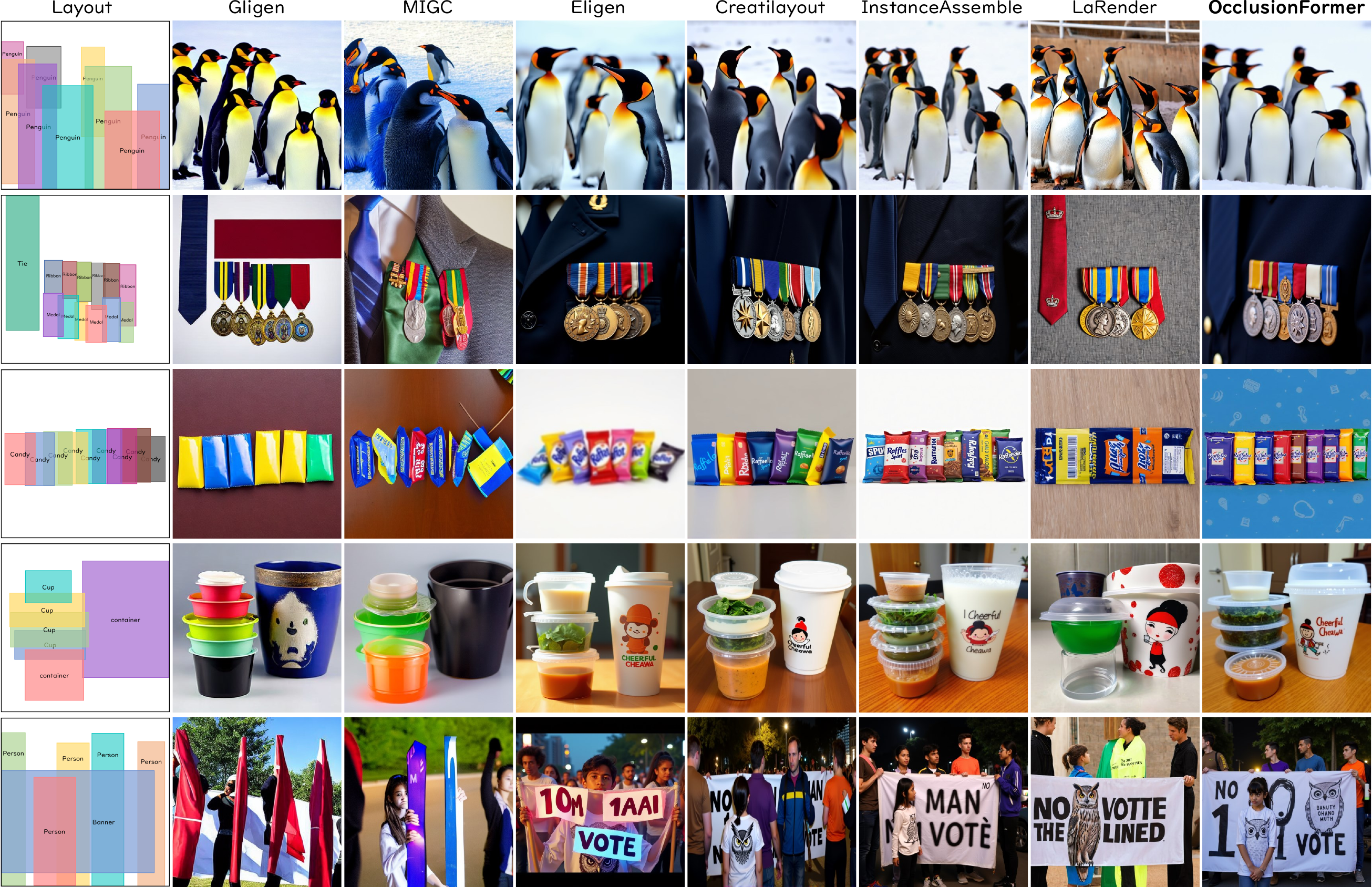}
        \caption{The visual comparison of different methods on our constructed \datasetname Eval.}
        \label{fig:supp_qualitative_results_overlaybench_real}
    \end{minipage}
\end{figure*}

\section{Investigation of Predicted Masks}
As visualized in \cref{fig:supp_mask_pred}, we investigate the evolution of the predicted foreground probability maps $\hat{\mathbf{M}}_i^{fg}$ predicted by mask predictor in the location of the first single-stream block throughout the denoising process.  At the early stages, the mask predictor focuses on capturing the coarse and amodal spatial footprint of the instances, roughly filling the entire bounding box area. However, as the denoising progresses toward the final steps, the masks gradually become sharper and conform to the fine-grained object boundaries. 

\vspace{-2mm}

\section{More Qualitative Comparison}

We provide additional visual comparisons to further substantiate the qualitative superiority of the proposed \methodname over existing state-of-the-art methods. We provide additional visual comparisons to evaluate the generation quality. \cref{fig:supp_qualitative_results_overlaybench} presents results on OverLayBench~\cite{li2025overlaybench}, demonstrating our method's superiority in handling dense overlaps and preserving instance boundaries compared to previous methods. \cref{fig:supp_qualitative_results_overlaybench_real} showcases the performance on our \datasetname Eval benchmark, verifying that our approach consistently maintains high realism and structural fidelity even in complex real-world scenarios.

\section{More Ablation Results}
\cref{tab:ablation_full} details the full ablation study across OverLayBench and \datasetname Eval. We include Flux.1-dev to establish a lower bound for spatial metrics (e.g., mIoU, O-mIoU, Occ., Dep.). Notably, Flux achieves high CLIP-G scores due to direct derivation from global prompts. Regarding fidelity, Flux naturally exhibits low FID on OverLayBench as the dataset itself is synthesized by Flux. However, its FID performance degrades on the real-world \datasetname Eval. Beyond this baseline, we observe distinct trends regarding specific components.

First, removing instance decoupling (\textit{w/o Inst. Decouple}) results in the most severe degradation. For instance, in the Complex subset, mIoU drops significantly from 0.6037 to 0.5177, and Occlusion accuracy (Occ.) falls from 0.7797 to 0.6109. This confirms that decoupling is foundational, essential for preventing feature entanglement and ensuring individual instances are generated with distinct identities.

Second, the significance of explicit Z-order modeling (\textit{w/o Occlusion Cond.}) exhibits a clear correlation with scene complexity. On the Simple subset where overlaps are minimal, the model performs comparably to the full method (mIoU 0.7385 vs. 0.7405). However, on Complex subsets and \datasetname Eval featuring dense and intricate overlaps, the lack of explicit Z-order leads to a notable drop in performance (e.g., Occ. drops roughly 5.3\% on Complex). This demonstrates that while implicit learning suffices for simple layouts, explicit volumetric rendering is indispensable for resolving intricate occlusion relationships.

Third, regarding the spatial alignment components, both \textit{Learned Sigma} and \textit{Queried Loss} prove critical for fine-grained spatial precision. Removing the learned density (\textit{w/o Learned Sigma}) harms the ability to modulate opacity dynamically, leading to a decreased O-mIoU. Similarly, removing the alignment loss (\textit{w/o Queried Loss}) compromises boundary precision, evidenced by the decline in SR$_\text{E}$ across all sets (e.g., 0.8158 to 0.8024 on \datasetname Eval).

Finally, training without amodal annotations (\textit{w/o Amodal Data}) fails to maintain the structural integrity of occluded objects. While achieving competitive FID scores, the degradation in O-mIoU, Occlusion Order (Occ.), and Depth Order (Dep.) across Complex subsets and \datasetname Eval highlights that amodal supervision provides vital geometric signals for learning correct occlusion dependencies.

\vspace{-0.2cm}

\begin{table}[t]
    \centering
    \caption{User study results comparing Occ., Layout Align, Local Fidelity, and Global Align, the higher is better.}
    \label{tab:user_study}
    \small
    \begin{tabular}{lcccc}
        \toprule
        Method & Occ.$\uparrow$ & \makecell{Layout \\ Align} $\uparrow$ & \makecell{Local \\ Fidelity}$\uparrow$ & \makecell{Global \\ Align}$\uparrow$ \\
        \midrule
        GLIGEN & 0.5486 & 0.6257 & 0.4838 & 0.5152 \\
        MIGC & 0.2086 & 0.1790 & 0.1676 & 0.3657 \\
        LaRender & 0.5838 & 0.5433 & 0.4824 & 0.2314 \\
        Eligen & 0.5390 & 0.5776 & 0.5543 & 0.6876 \\
        Creatilayout & \underline{0.6743} & 0.6567 & 0.7295 & 0.7095 \\
        InstanceAssemble & 0.6424 & \underline{0.6919} & \underline{0.7738} & \underline{0.7390} \\
        \textbf{Ours} & \textbf{0.7833} & \textbf{0.7357} & \textbf{0.8086} & \textbf{0.7514} \\
        \bottomrule
    \end{tabular}
    \vspace{-6mm}
\end{table}

\section{User Study}
We conducted a user study employed 15 participants with 300 randomly selected samples from the OverLayBench \textit{Complex} subset and our constructed \textit{\datasetname Eval} benchmark. 
Evaluators ranked the images of 7 methods based on occlusion accuracy, layout alignment, local fidelity, and global alignment. Scores were assigned from 1 to 7 based on the ranking and normalized to $[1/7, 1]$. 
As reported in \cref{tab:user_study}, our method achieves the highest ratings across all four dimensions, confirming its superiority in human perceptual evaluation over existing state-of-the-art baselines.
\vspace{-3mm}

\begin{figure}[H] 
    \centering
    \includegraphics[width=\linewidth]{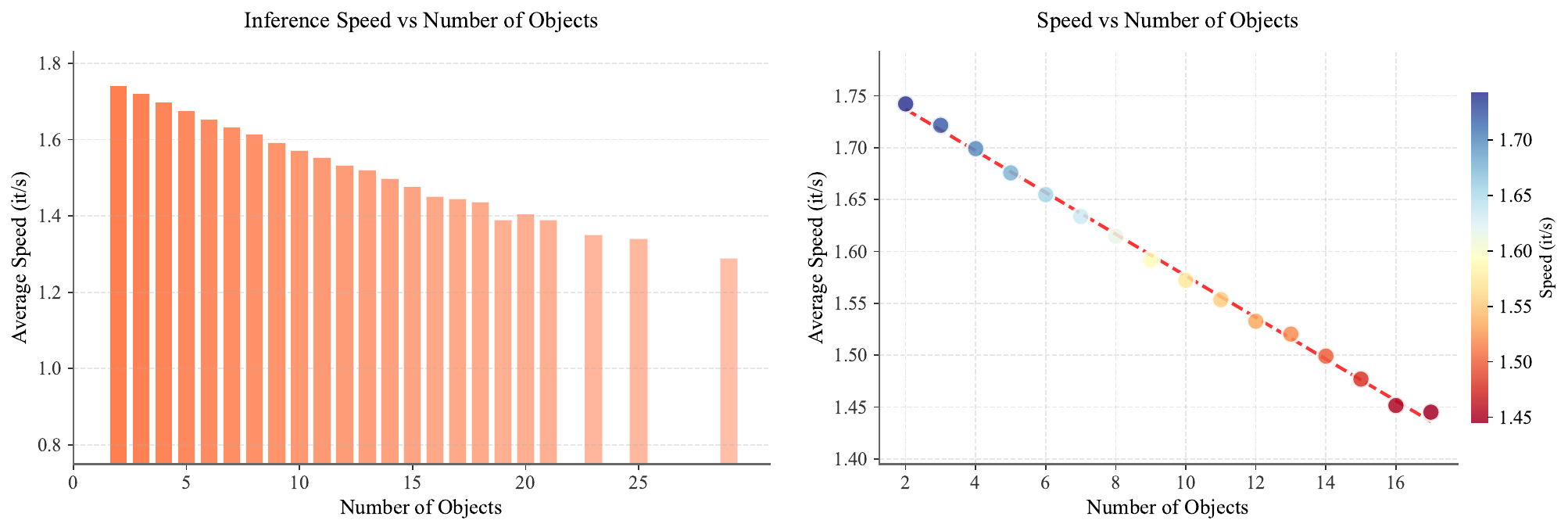}
    \caption{\textbf{Efficiency analysis.} We report the inference speed on NVIDIA A800 GPU with varying numbers of objects. The results show a linear scaling trend, ensuring efficiency in dense scenes.}
    \label{fig:speed}
    
    \vspace{-6mm}
\end{figure}

\section{Efficiency Analysis}

We investigate the computational efficiency of our proposed framework by evaluating the inference speed on a single NVIDIA A800 GPU. Given that our method employs an instance decoupling strategy to process local features, the computational cost is correlated with the scene complexity. As illustrated in \cref{fig:speed}, we observe a linear relationship between the number of objects and the generation speed. Although the inference time naturally increases as the scene becomes more cluttered, the decline in speed remains gradual and stable. This demonstrates that our approach scales effectively and maintains practical efficiency even when handling scenarios with a large number of instances.

\vspace{-0.3cm}

\begin{figure*}[t]
    \centering
    \includegraphics[width=0.85\linewidth]{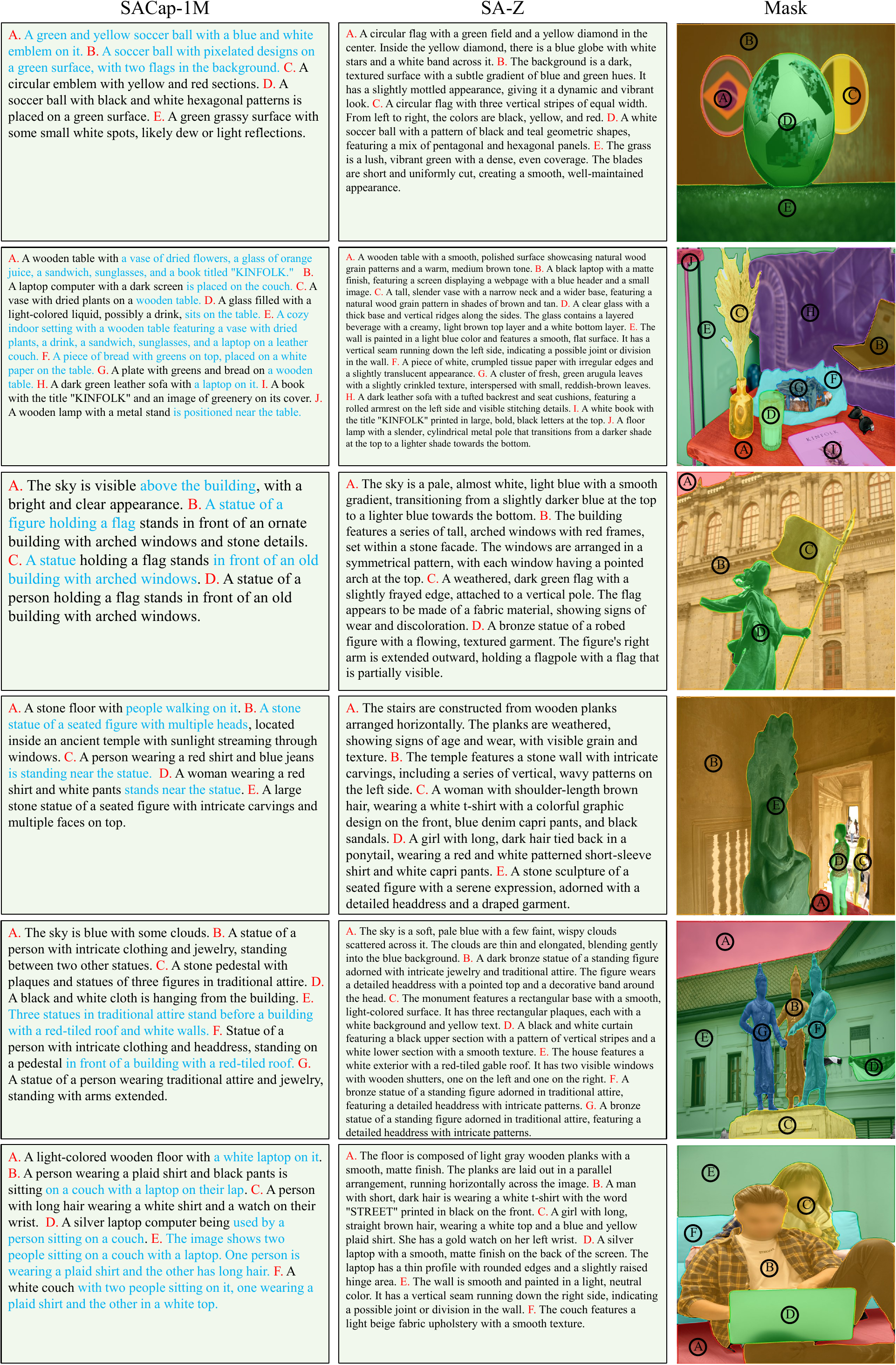}
    \caption{The comparison of captions between SACap-1M~\cite{liseg2any} and \datasetname (Ours).}
    \vspace{-0.4cm}
    \label{fig:dataset_comparison}
\end{figure*}

\section{Illustration of Noise in SACap-1M}
The comparison in \cref{fig:dataset_comparison} illustrates the annotation noise inherent in the SACap-1M~\cite{liseg2any} dataset where images are resize to 1:1 for better view. SACap-1M generates regional captions by prompting the Qwen2-VL-72B~\cite{bai2025qwen2} model with bounding box coordinates. However, this box-based prompting mechanism inevitably introduces noise, as rectangular bounding boxes rarely align perfectly with irregular object shapes. Consequently, the boxes often encompass background elements or adjacent instances, leading the VLM to erroneously attribute these surrounding visual features to the target entity. 

To address this limitation, we utilize DescribeAnything~\cite{lian2025describe} to perform precise mask-level annotation. By constraining the visual analysis strictly to the segmented regions, our approach effectively filters out context-induced noise (highlighted by the \textcolor[RGB]{0, 176, 240}{blue marks} in the \cref{fig:dataset_comparison}). As a result, \datasetname derives significantly cleaner and more detailed captions that strictly adhere to the visual attributes of the specific instances of interest.

\vspace{-2mm}

\begin{figure*}[t]
    \centering
    \includegraphics[width=0.9\linewidth]{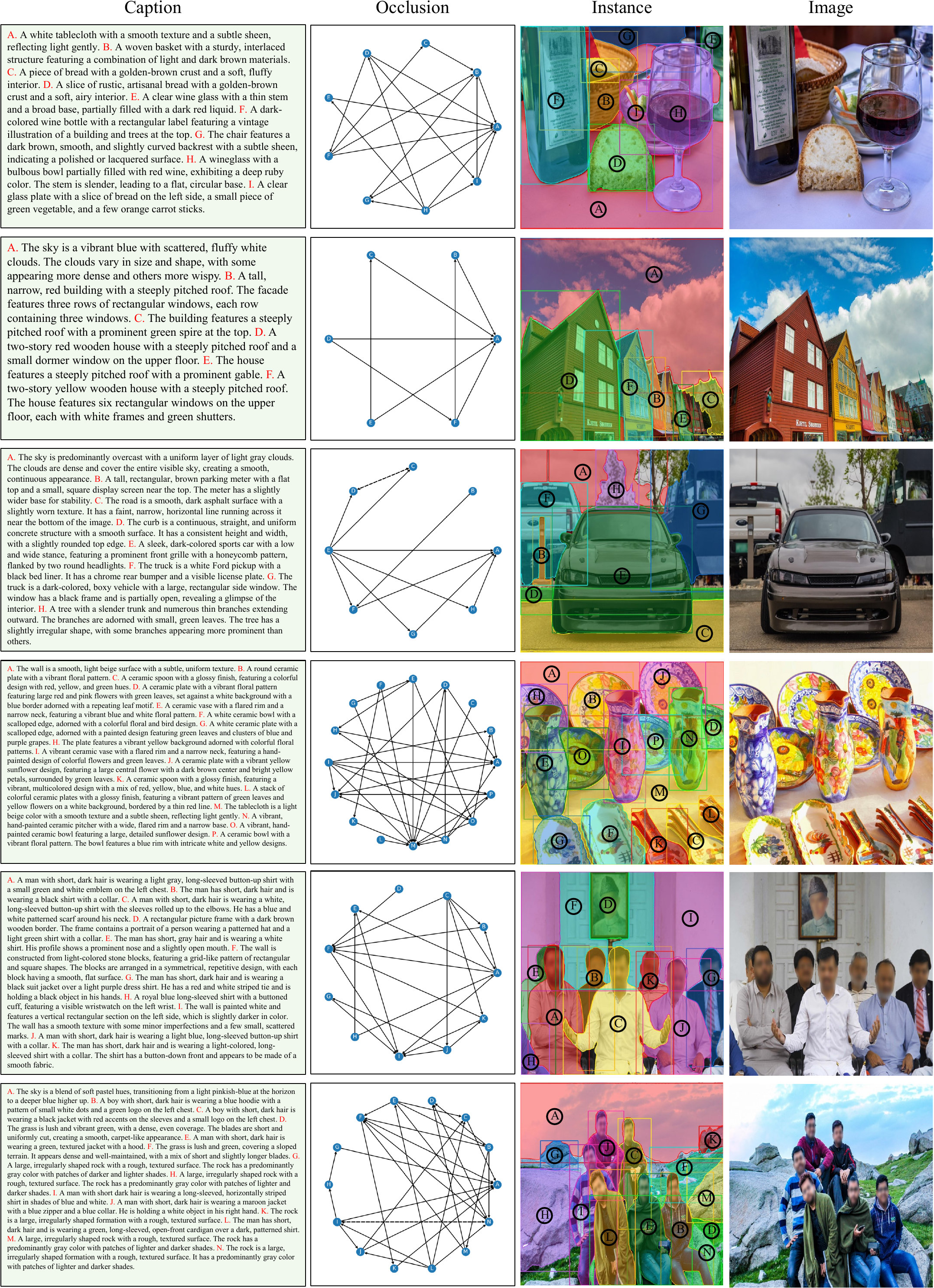}
    \caption{Examples from \datasetname, where arrows in the occlusion graphs denote the ``occludes" relationship.}
    \vspace{-0.4cm}
    \label{fig:SA-Z}
\end{figure*}

\section{Examples of \datasetname}
Figure \ref{fig:SA-Z} provides examples of the training dataset sampled from \datasetname. Our dataset provides detailed annotations for mask areas and pairwise instance occlusion relationships. We also incorporate SAM-3D~\cite{chen2025sam} to extract the amodal annotations for the occluded instance.

\vspace{-2.5mm}

\begin{figure*}[t]
    \centering
    \includegraphics[width=0.9\linewidth]{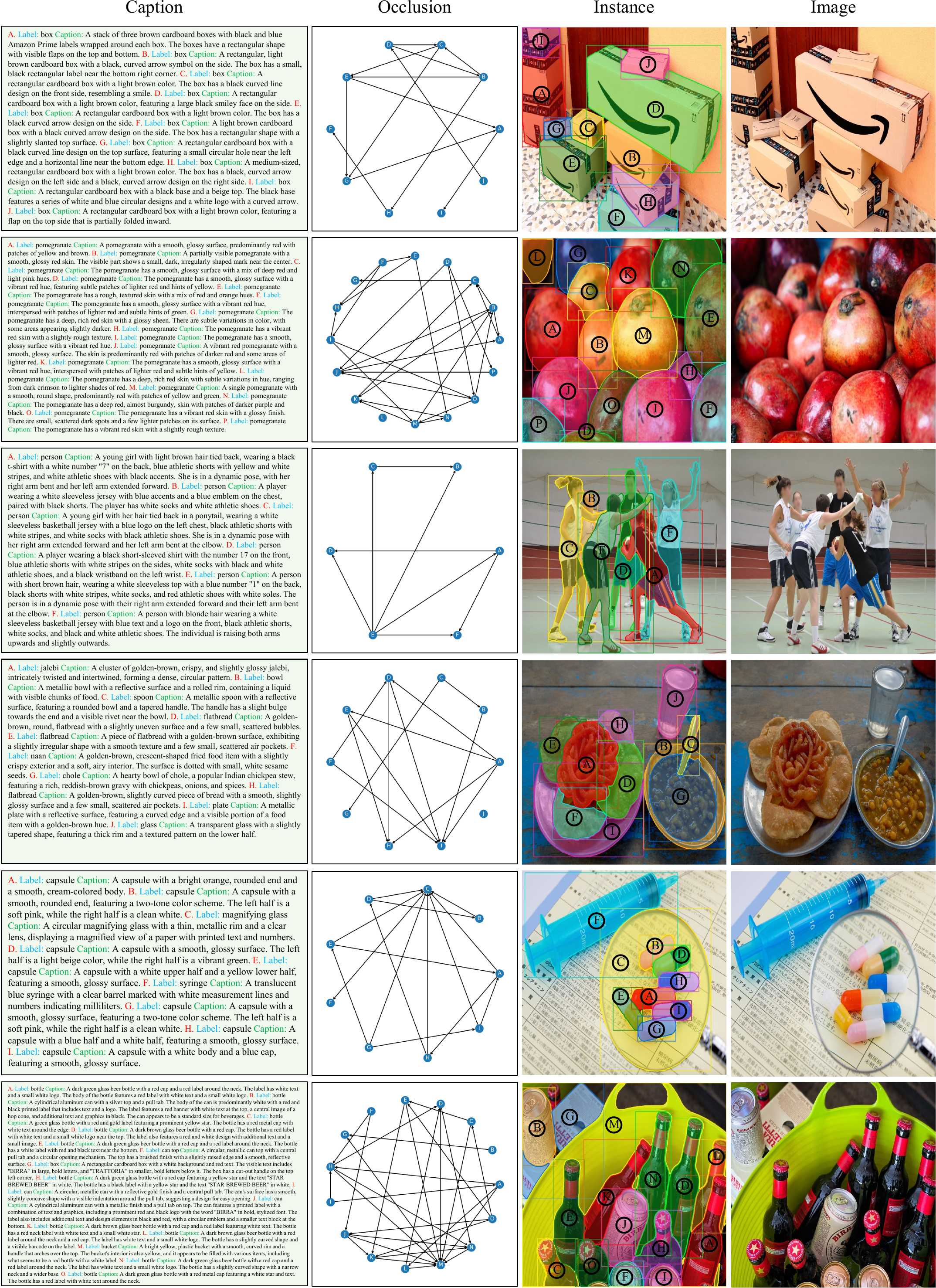}
    \caption{Examples from our created \datasetname Eval, where arrows in the occlusion graphs denote the ``occludes" relationship.}
    \vspace{-0.4cm}
    \label{fig:benchmark_example}
\end{figure*}

\section{Examples and Statistics of the \datasetname Eval}
\vspace{-0.2cm}
\begin{figure}[H] 
    \centering
    \includegraphics[width=\linewidth]{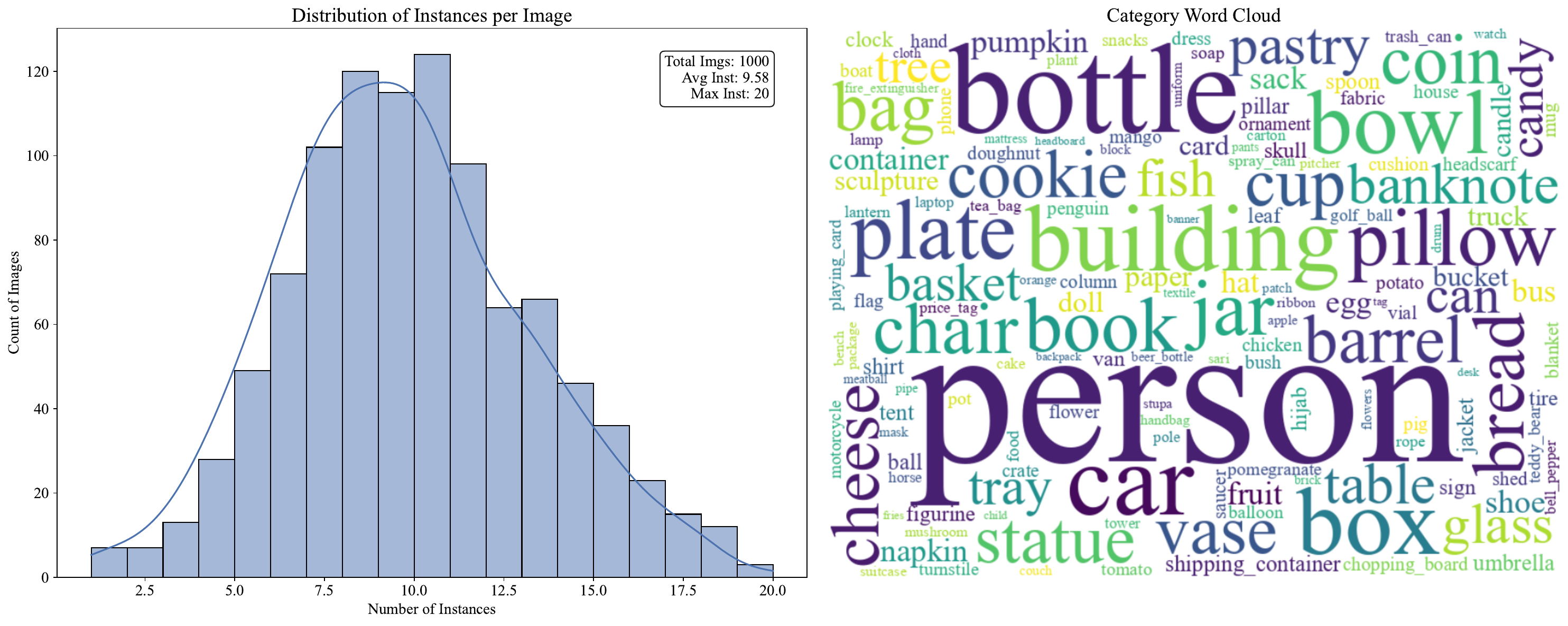}
    \caption{Statistical overview of \datasetname Eval. The left shows the distribution of instances per image, while the right word cloud illustrates the semantic diversity across 749 categories.}
    \label{fig:benchmark_statistic}
    \vspace{-4mm}
\end{figure}

\cref{fig:benchmark_statistic} presents key statistics of our \datasetname Eval benchmark. As shown in the statistical plots, the benchmark encompasses 749 distinct categories with varying instance densities per image, ensuring both semantic breadth and scene complexity. To ensure consistency with OverLayBench~\cite{li2025overlaybench}, we employ Qwen-VL-32B~\cite{bai2025qwen2} to generate semantic labels and filter out non-salient objects from SA-1B~\cite{kirillov2023segment}. The examples are in \cref{fig:benchmark_example}. Notably, we adopt modal bounding boxes for evaluation instead of amodal ones to minimize ambiguity. Since amodal boxes encompass occluded regions that lack corresponding visual pixels, using them for spatial metrics like IoU would introduce misalignment with the generated content, whereas modal boxes provide a more grounded reference for evaluating visual layout accuracy.

\end{document}